\documentclass[letterpaper]{article} 
\usepackage[preprint]{aaai2027}  
\usepackage[hyphens]{url}  
\usepackage{graphicx} 
\usepackage{natbib}  
\usepackage{caption} 
\usepackage{algorithm}
\usepackage{algorithmic}
\usepackage{multirow}
\usepackage{amsfonts}
\usepackage{amsmath}

\usepackage{newfloat}
\usepackage{listings}
\DeclareCaptionStyle{ruled}{labelfont=normalfont,labelsep=colon,strut=off} 
\floatstyle{ruled}
\newfloat{listing}{tb}{lst}{}
\floatname{listing}{Listing}

\usepackage{booktabs}

\title{Do Pathology Vision-Language Models Truly See Pathology?}
\author{
    Chengyang Zhang\textsuperscript{\rm 1,2},
    Wenchuang Zhang\textsuperscript{\rm 2},
    Bo Li\textsuperscript{\rm 3},
    Xinyu Liu\textsuperscript{\rm 1},
    Jiaming Yang\textsuperscript{\rm 1},
    Mengran Li\textsuperscript{\rm 4},
    Chenxun Deng\textsuperscript{\rm 5},
    Jie Chen\textsuperscript{\rm 2},
    Yang Zhang\textsuperscript{\rm 3},
    Wei Ju\textsuperscript{\rm 1},
    Yuhao Yi\textsuperscript{\rm 1,2}\corresponding,
    Hong Bu\textsuperscript{\rm 2},
    Jiancheng Lv\textsuperscript{\rm 1}
}
\affiliations{
    \textsuperscript{\rm 1}College of Computer Science, Sichuan University\\
    \textsuperscript{\rm 2}Department of Pathology and Institute of Clinical Pathology, West China Hospital, Sichuan University\\
    \textsuperscript{\rm 3}Department of Computer Science, School of Computing, National University of Singapore\\
    \textsuperscript{\rm 4}School of Intelligent Systems Engineering, Sun Yat-sen University\\
    \textsuperscript{\rm 5}Institute of Automation, Chinese Academy of Sciences\\

    yuhaoyi@scu.edu.cn
}

\begin{document}

\maketitle

\begin{abstract}
Pathology vision-language models (VLMs) have recently progressed rapidly and are commonly evaluated by answer accuracy on pathology VQA benchmarks. However, we dig into current evaluations and identify three overlooked issues: 1) Visual evidence is not always necessary. For instance, Gemini-3-Pro achieves 53.5\% average accuracy across 5 VQA benchmarks without any visual input. 2) Domain training can improve accuracy without proportional gains in visual binding. Compared with Qwen2.5-VL-7B, Patho-R1-7B exhibits a 5.8-point lower multimodal gain and a 3.7-point lower attention IoU. 3) Entity-level attention is diffuse and weakly query-specific. On PathVG, attention maps remain highly correlated across different entity queries. These issues can lead to substantial misjudgments of pathology VLMs' actual multimodal capabilities. To this end, we present \textbf{PathBind}, a benchmark comprising 2,600 samples: PathBind-VQA with 1,500 questions across six dimensions, PathBind-PTA with 600 questions from a private pathology teaching atlas, and PathBind-Grounding with 500 expert-curated region-level samples. Each component undergoes task-specific automated filtering and expert review to reduce textual shortcuts and improve entity-region correspondence. We evaluate 18 representative VLMs on VQA samples of PathBind and five existing pathology VQA benchmarks, and further evaluate 10 VLMs on PathBind-Grounding and PathVG. Results show that current pathology VLMs still exhibit a substantial gap between answer-side performance and visual-semantic binding.
\end{abstract}


\section{Introduction}

Pathology vision-language models (VLMs) have rapidly advanced through pathology-specific pretraining or post-training, achieving strong performance on pathology visual question answering (VQA) benchmarks \cite{sun2024pathmmu,he2020pathvqa,zuo2025medxpertqa,quilt1m,hu2024omnimedvqa}. Answer accuracy has consequently become a standard metric for pathology understanding. However, accuracy alone does not reveal whether a model uses the histopathology image, relies on pathology-specific language priors, or associates the queried concept with the relevant microscopic evidence. This raises a fundamental question: \emph{do pathology VLMs truly see pathology, or do they primarily become better pathology answerers?}

\begin{figure}[t]
  \includegraphics[width=1.\columnwidth]{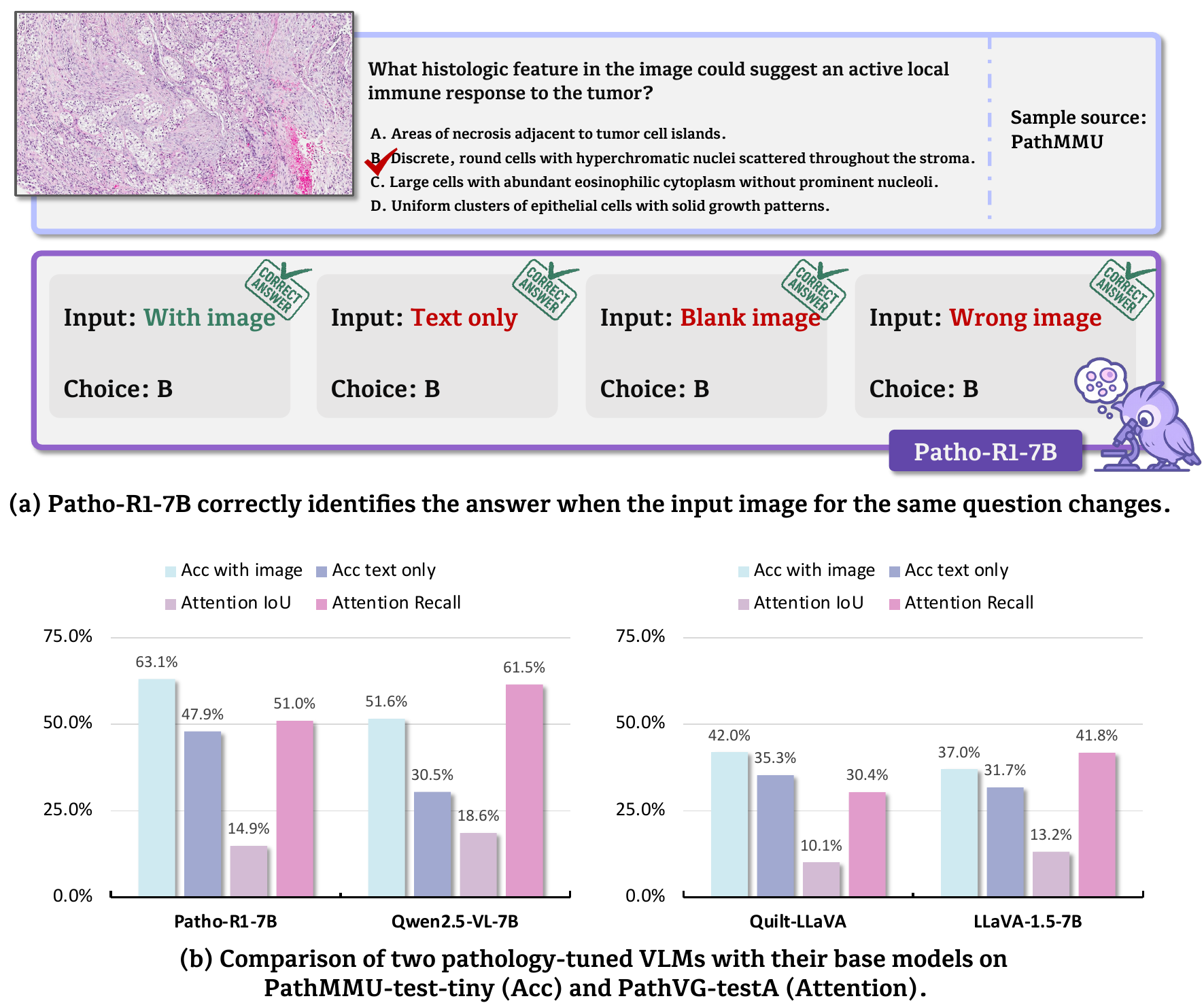}
  \caption{(a) Pathology-tuned VLM answering correctly under text-only, blank-image, or wrong-image conditions. (b) Pathology-specific training improves answer accuracy but does not proportionally improve multimodal gain or visual grounding.}
  \label{fig1}
\end{figure}

We argue that pathology understanding requires more than selecting the correct answer. Pathologists justify conclusions by locating and interpreting visual evidence, including glandular architecture, lymphocytic infiltration, tumor nests, and other cellular or tissue-level patterns. For questions that depend on localized microscopic evidence, \emph{a pathology VLM should therefore not only predict the correct answer, but also associate the queried pathological concept with the supporting image region.} Without evaluating this evidence chain, benchmark accuracy may conflate genuine visual reasoning with textual shortcuts and domain knowledge.


Figure \ref{fig1} illustrates this gap. In a representative PathMMU example, Patho-R1-7B selects the same correct option when given the original image, no image, a blank image, or an unrelated image, suggesting that this prediction may be dominated by non-visual cues. This behavior extends beyond individual examples. Compared with Qwen2.5-VL-7B, Patho-R1-7B achieves higher image-conditioned accuracy, but its text-only accuracy increases even more, resulting in lower multimodal gain and weaker attention-based grounding. A similar pattern appears for Quilt-LLaVA relative to LLaVA-1.5-7B. We refer to this discrepancy as the domain training illusion: pathology-specific training improves answer accuracy without proportional gains in visual binding.


Existing pathology VQA protocols are not designed to expose this discrepancy. They typically provide an image-question pair, infer an answer, and report accuracy, without testing whether the image is necessary, whether predictions are sensitive to visual perturbations, or whether entity-token attention aligns with the annotated pathological region. As a result, models may achieve higher scores by exploiting question-option associations and pathology language priors while remaining weakly grounded in microscopic evidence. A more faithful evaluation should therefore distinguish three complementary capabilities: visual dependence, multimodal gains introduced by domain-specific training, and entity-level visual grounding.



To support such evaluation, we introduce \textbf{PathBind}, a benchmark designed to evaluate visual-semantic binding. PathBind consists of three complementary components. \textbf{PathBind-VQA} includes 1,500 questions distilled from five public pathology benchmarks through text-only filtering and expert review. \textbf{PathBind-PTA} contains 600 questions derived from a private pathology teaching atlas, and \textbf{PathBind-Grounding} provides 500 expert-curated region-level samples. We evaluate 18 representative VLMs on VQA samples, and conduct attention-based grounding analysis on 10 VLMs with accessible attention weights. The experimental results indicate a substantial gap between answer correctness and entity-level visual grounding. In summary, our contributions are threefold:
\begin{itemize}
    \item We identify three issues of current pathology VLM evaluation: (1) visual evidence is not always necessary in existing benchmarks; (2) domain-specific training improves answer accuracy
    without proportional gains in visual binding; and (3) entity-level attention remains diffuse and weakly query-specific.
    \item We construct PathBind, a diagnostic benchmark of 2,600 samples spanning three components: PathBind-VQA, PathBind-PTA, and PathBind-Grounding. Each part is filtered by automated pipelines and expert review.
    \item We benchmark 18 representative VLMs and show that strong answer-side performance does not imply correspondingly strong visual-semantic binding. 
\end{itemize}

\section{Related Works}
\paragraph{Pathology vision-language models.} 
As large vision-language models continue to advance, a growing body of work has adapted vision-language learning to computational pathology. Central to this direction are early pathology vision-language foundation models that transfer contrastive image-text alignment to histopathology, such as CONCH, which establish pathology-specific visual-language representations from large-scale image-text pairs and captions \cite{lu2024visual}. Built upon such modality-alignment foundations and general LVLM architectures, recent pathology VLMs further connect pathology image encoders with language models through domain-specific pretraining, instruction tuning, multi-resolution modeling, whole-slide modeling, and reasoning-oriented post-training \cite{seyfioglu2024quilt,sun2025pathgen,sun2025cpath,ding2025multimodal,sun2024pathasst,zhang2026patho,wu2025pathvlm,chen2024wsi}. For example, Quilt-LLaVA \cite{seyfioglu2024quilt} and PathGen-LLaVA \cite{sun2025pathgen} represent instruction-tuned pathology assistants, CPath-Omni \cite{sun2025cpath} extends multimodal modeling toward WSI-level analysis, and Patho-R1 \cite{zhang2026patho} explores reinforcement learning-based pathology reasoning. Recent works \cite{jiang2026pathreasoner,ghezloo2025pathfinder,chen2025pathagent,sun2026cpathagent,zhang2026pathoagentic,lyu2026wsi} further improve multimodal data quality and extend single-VLM systems toward agentic reasoning, enabling more flexible and powerful diagnostic workflows.

\paragraph{Evaluations of pathology VLMs.}
To probe the capabilities of pathology VLMs, the community has developed pathology-specific VQA datasets, expert-level multimodal benchmarks, model evaluation suites, and grounding datasets \cite{sun2024pathmmu,he2020pathvqa,zuo2025medxpertqa,quilt1m,hu2024omnimedvqa,gilal2025pathvlm,sun2025pathbench,chen2026omnipathovqa,majzoub2025good}. Early pathology VQA datasets evaluate whether models can answer questions about pathology images, while recent benchmarks expand evaluation to multimodal reasoning across diverse pathology sources and subdomains. For example, PathMMU \cite{sun2024pathmmu} comprises over 33K multimodal multiple-choice questions and 24K pathology images. Meanwhile, pathology visual grounding benchmarks such as PathVG \cite{zhong2025pathvg} further extend evaluation from question answering to region localization, enabling region-level assessment of pathological expressions. However, existing evaluations of pathology VLMs overlook several critical issues. On the one hand, they do not guarantee that all evaluation samples require visual evidence to be correctly answered, leaving room for models to exploit textual priors or dataset-specific shortcuts. On the other hand, current evaluations mainly follow the standard protocol of inferring on fixed benchmarks and reporting answer accuracy, without examining whether pathology training improves visual-semantic binding between pathological concepts and microscopic regions. This oversight can lead to an overestimation of pathology understanding and misjudgments of the real visual grounding gains achieved by domain-specific multimodal training.

\begin{table*}[t!]
\renewcommand{\arraystretch}{1.2}
\renewcommand{\tabcolsep}{8pt}
\centering
\resizebox{\textwidth}{!}
{\begin{tabular}{l|l|cccccc|c}
\toprule[1pt]
\multirow{2}{*}{Model}               & \multirow{2}{*}{Strategy} & \multicolumn{2}{c}{PathMMU} & \multicolumn{2}{c}{YorN} & MedXpert & OmniMed & \multirow{2}{*}{Avg.} \\ \cline{3-8} 
                                     &                           & Test-tiny     & Test-all    & Quilt-VQA   & Path-VQA   & Path       & Bright     \\ \midrule[0.7pt]
\multicolumn{9}{c}{\textit{Baseline}}                                                                                                                        \\ \midrule[0.7pt]
Random Choice                        & -                         & 23.4          & 24.2        & 50.0        & 50.0       & 21.3       & 24.2   & 32.2    \\ \midrule[0.7pt]
\multicolumn{9}{c}{\textit{Closed-source VLMs}}                                                                                                             \\ \midrule[0.7pt]
\multirow{2}{*}{GPT-5.4}             & VLM                & 70.8\textcolor{black}{$_{\uparrow 25.1}$}              & 68.6\textcolor{black}{$_{\uparrow 21.6}$}           & 68.2\textcolor{black}{$_{\uparrow 25.7}$}            & 71.1\textcolor{black}{$_{\uparrow 23.4}$}           & 68.9\textcolor{black}{$_{\uparrow 14.4}$}           & 61.2\textcolor{black}{$_{\uparrow 17.2}$}   & 68.1\textcolor{black}{$_{\uparrow 21.2}$}        \\
                                     & VLM-text                 & 45.7              & 47.0            & 42.5            & 47.7           & \underline{54.5}           & 44.0   & 46.9       \\
\multirow{2}{*}{Gemini-3-Pro}        & VLM                &\textbf{77.8}\textcolor{black}{$_{\uparrow 26.5}$}            & \textbf{75.6}\textcolor{black}{$_{\uparrow 25.1}$}            & \textbf{73.5}\textcolor{black}{$_{\uparrow 24.8}$}            & \textbf{77.5}\textcolor{black}{$_{\uparrow 17.8}$}           & \textbf{80.0}\textcolor{black}{$_{\uparrow 25.6}$}           & \textbf{71.8}\textcolor{black}{$_{\uparrow 15.7}$}     &  \textbf{76.0}\textcolor{black}{$_{\uparrow 22.5}$}    \\
                                     & VLM-text                 & \underline{51.3}              & \underline{50.5}            & 48.7            & \underline{59.7}           & 54.4           & 56.1     & \underline{53.5}     \\ \midrule[0.7pt]
\multicolumn{9}{c}{\textit{Open-source general VLMs and pathology VLMs}}                                                                                     \\ \midrule[0.7pt]
\multirow{2}{*}{LLaVA-1.5-7B}        & VLM                &  37.0\textcolor{black}{$_{\uparrow 5.3}$}             & 36.1\textcolor{black}{$_{\uparrow 5.4}$}            & 56.6\textcolor{black}{$_{\uparrow 14.6}$}            & 49.5\textcolor{black}{$_{\uparrow 9.9}$}           & 22.2\textcolor{black}{$_{\uparrow 7.8}$}           & 46.5\textcolor{black}{$_{\uparrow 12.3}$}      & 41.3\textcolor{black}{$_{\uparrow 9.2}$}    \\
                                     & VLM-text                 & 31.7              & 30.7            & 42.0            & 39.6           & 14.4           & 34.2   & 32.1        \\
\multirow{2}{*}{Quilt-LLaVA}         & VLM                &  42.0\textcolor{black}{$_{\uparrow 6.7}$}             & 42.2\textcolor{black}{$_{\uparrow 5.4}$}            & 64.3\textcolor{black}{$_{\uparrow 10.3}$}            & 57.3\textcolor{black}{$_{\uparrow 1.1}$}           & 21.1\textcolor{black}{$_{\uparrow 4.4}$}           & 49.8\textcolor{black}{$_{\uparrow 10.6}$}        & 46.1\textcolor{black}{$_{\uparrow 6.4}$}    \\
                                     & VLM-text                 & 35.3              & 36.8            & \underline{54.0}            & 56.2           & 16.7           & 39.2     & 39.7     \\
\multirow{2}{*}{LLaVA-1.5-13B}       & VLM                &  36.3\textcolor{black}{$_{\uparrow 4.2}$}             & 37.2\textcolor{black}{$_{\uparrow 5.1}$}            &  58.3\textcolor{black}{$_{\uparrow 14.1}$}           & 60.8\textcolor{black}{$_{\uparrow 13.4}$}           & 18.9\textcolor{black}{$_{\uparrow 4.6}$}           & 45.2\textcolor{black}{$_{\uparrow 11.9}$}         & 42.8\textcolor{black}{$_{\uparrow 8.9}$} \\
                                     & VLM-text                 & 32.1              & 32.1            & 44.2            & 47.4           & 14.3           & 33.3     & 33.9     \\
\multirow{2}{*}{PathGen-LLaVA}       & VLM                &  62.6\textcolor{black}{$_{\uparrow 11.4}$}             & 60.4\textcolor{black}{$_{\uparrow 11.3}$}            & 62.3\textcolor{black}{$_{\uparrow 13.7}$}            & 63.1\textcolor{black}{$_{\uparrow 4.9}$}           & 30.0\textcolor{black}{$_{\uparrow 6.7}$}           & 68.5\textcolor{black}{$_{\uparrow 11.2}$}         & 57.8\textcolor{black}{$_{\uparrow 9.8}$} \\
                                     & VLM-text                 & 51.2              & 49.1            & 48.6            & 58.2           & 23.3           & 57.3    & 48.0      \\
\multirow{2}{*}{Qwen2-VL-7B}         & VLM                & 48.6\textcolor{black}{$_{\uparrow 13.5}$}              & 47.0\textcolor{black}{$_{\uparrow 12.3}$}            & 52.6\textcolor{black}{$_{\uparrow 14.3}$}            & 60.6\textcolor{black}{$_{\uparrow 14.1}$}           & 18.9\textcolor{black}{$_{\uparrow 5.6}$}           & 51.9\textcolor{black}{$_{\uparrow 9.0}$}          & 46.6\textcolor{black}{$_{\uparrow 11.5}$} \\
                                     & VLM-text                 & 35.1              & 34.7            & 38.3            & 46.5           & 13.3           & 42.9    & 35.1      \\
\multirow{2}{*}{HuatuoGPT-Vision-7B} & VLM                & 57.5\textcolor{black}{$_{\uparrow 11.8}$}              & 56.1\textcolor{black}{$_{\uparrow 12.0}$}            & 62.6\textcolor{black}{$_{\uparrow 15.4}$}            & 60.2\textcolor{black}{$_{\uparrow 10.7}$}           & 20.0\textcolor{black}{$_{\uparrow 2.2}$}           & 44.9\textcolor{black}{$_{\uparrow 7.1}$}         & 50.2\textcolor{black}{$_{\uparrow 9.8}$} \\
                                     & VLM-text                 & 45.7              & 44.1            & 47.2            & 49.5           & 17.8           & 37.8     & 40.4     \\
\multirow{2}{*}{Qwen2.5-VL-7B}       & VLM                & 51.6\textcolor{black}{$_{\uparrow 21.1}$}              & 48.7\textcolor{black}{$_{\uparrow 17.7}$}            &  54.2\textcolor{black}{$_{\uparrow 14.1}$}           & 59.5\textcolor{black}{$_{\uparrow 11.7}$}           & 13.3\textcolor{black}{$_{\uparrow 1.1}$}           & 54.6\textcolor{black}{$_{\uparrow 9.9}$}         & 47.0\textcolor{black}{$_{\uparrow 12.6}$} \\
                                     & VLM-text                 & 30.5              & 31.0            & 40.1            & 47.8           & 12.2           & 44.7   & 34.4       \\
\multirow{2}{*}{Patho-R1-7B}         & VLM                &  63.1\textcolor{black}{$_{\uparrow 15.3}$}             & 62.9\textcolor{black}{$_{\uparrow 18.3}$}            &  64.4\textcolor{black}{$_{\uparrow 11.6}$}           & 55.3\textcolor{black}{$_{\uparrow 12.3}$}           & 16.7\textcolor{black}{$_{\uparrow 0.9}$}           & 69.9\textcolor{black}{$_{\uparrow 11.3}$}        & 55.4\textcolor{black}{$_{\uparrow 11.6}$}   \\
                                     & VLM-text                 & 47.8              & 44.6            & 52.8            & 43.0           & 15.8           & \underline{58.6}    & 43.8      \\ \bottomrule[1pt]
\end{tabular}
}
\caption{Evaluations of various VLMs on five popular pathology VQA benchmarks. For the "strategy" column, "VLM" denotes evaluating with images, while "VLM-text" means evaluating without images. We report the results for 2 closed-source VLMs and 4 pathology-tuned VLMs, along with their base models. For additional open-source VLMs and the complete model list, please refer to Appendix \ref{appA1}. The highest results of the "VLM" and "VLM-text" settings across models are highlighted in \textbf{bold} and \underline{underlined}. The numbers with arrows indicate the improvement of “VLM” relative to “VLM-text”.}
\label{table1}
\end{table*}

\section{Diagnosing Pathology VLMs: Three Overlooked Issues}
\label{sec3}
In this section, we examine three overlooked issues in the current evaluation of pathology VLMs. We present detailed experimental results on existing pathology benchmarks and grounding data to substantiate our observations.

\subsection{First Issue: Visual Evidence Is Not Always Necessary}
A qualified pathology VQA sample should require the model to inspect and reason over the histopathology image. Otherwise, the evaluation may degenerate into testing the language backbone's textual and domain-prior capabilities. To examine this issue, we evaluate pathology VLMs, their base models, and representative general VLMs in two settings across existing pathology VQA benchmarks: the standard VLM setting, where the image is provided, and the VLM-text setting, where the model answers the same question without visual input.

As shown in Table \ref{table1}, many models retain substantial accuracy without images. Closed-source VLMs achieve strong text-only performance, with GPT-5.4 and Gemini-3-Pro reaching average VLM-text accuracies of 46.9\% and 53.5\%, respectively. Pathology-tuned open-source models show a similar pattern. PathGen-LLaVA obtains 48.0\% average accuracy without images, and Patho-R1 reaches 43.8\%. These scores are far above the random-choice baseline, indicating that a considerable portion of pathology VQA performance can be explained by question-option priors and pathology-domain language knowledge.

The gap between image-conditioned and text-only accuracy also varies across models and datasets. For example, Quilt-LLaVA improves by 6.4 points on average when images are added, and PathGen-LLaVA improves by 9.8 points. Even for stronger models, high image-conditioned accuracy is often accompanied by high text-only accuracy. These observations do not imply that images are useless for pathology VQA, but show that answer accuracy contains a substantial text-prior. Therefore, accuracy alone cannot faithfully measure whether a pathology VLM depends on visual evidence.

\begin{figure}[t!]
  \includegraphics[width=1.\columnwidth]{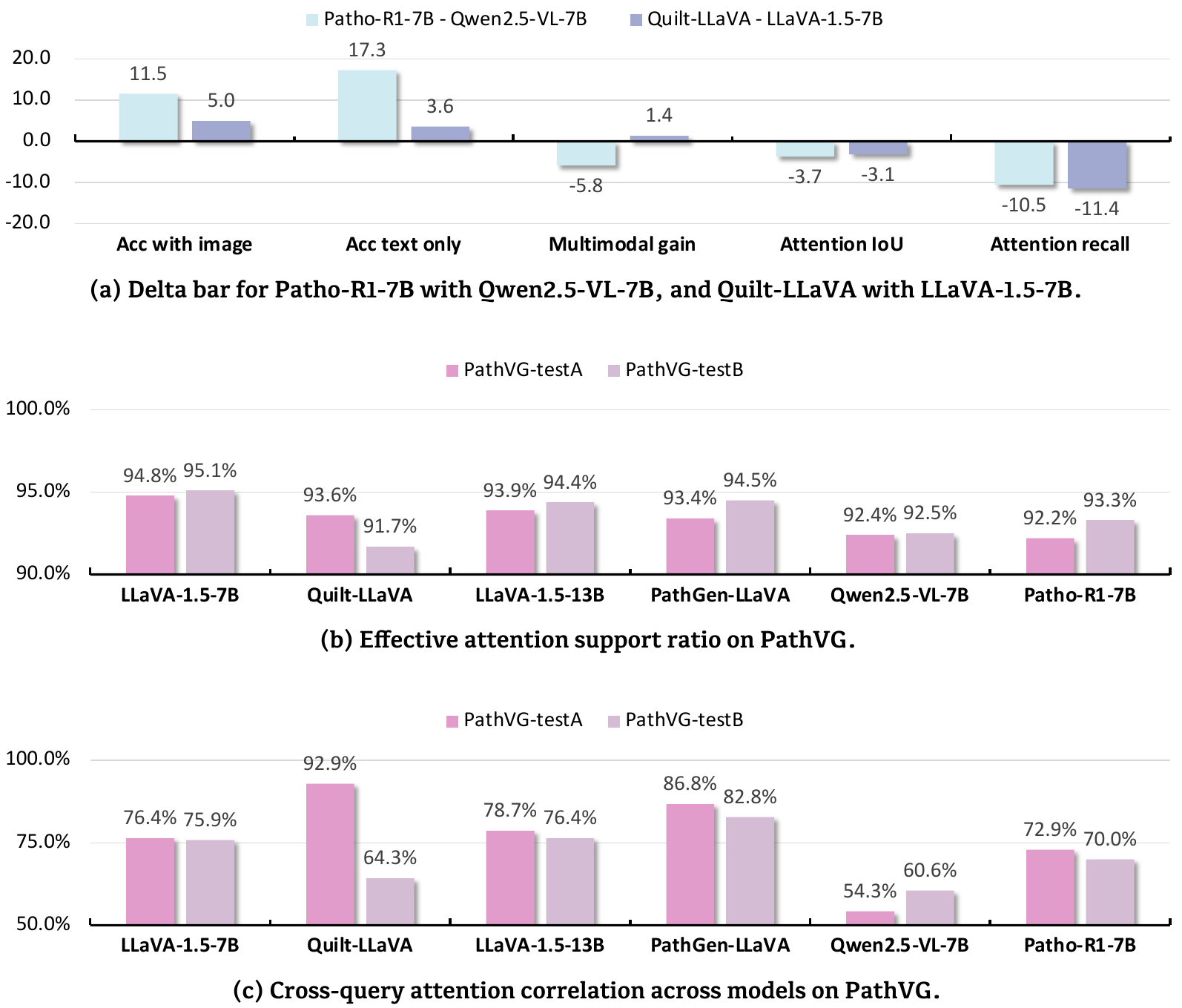}
  \caption{(a) Paired deltas on PathMMU-test-tiny for answer accuracy and multimodal gain, and on PathVG-testA for attention IoU and recall. (b) Entity-token attention exhibits near-full entropy-equivalent support on PathVG. (c) Attention maps for different pathological entity queries remain highly correlated.}
  \label{fig2}
\end{figure}

\subsection{Second Issue: Domain Training Can Improve Accuracy Without Improving Visual Binding}
Pathology-specific training is expected to improve visual pathology reasoning, not merely textual plausibility. However, paired comparisons between pathology-tuned models and their base models reveal a mismatch between answer improvement and grounding improvement. We refer to this mismatch as the \emph{Domain Training Illusion}: domain-specific training improves benchmark accuracy, but the gain is not accompanied by proportional improvement in visual binding, including behavioral visual dependence and entity-level grounding.

Figure \ref{fig2} (a) provides two representative comparisons. Patho-R1 improves over Qwen2.5-VL by 11.5 points in image-conditioned accuracy and by 17.3 points in text-only accuracy. However, its multimodal gain decreases by 5.8 points, and its attention IoU and attention recall decrease by 3.7 points and 10.5 points, respectively. A similar pattern appears in the LLaVA family: Quilt-LLaVA improves over LLaVA-1.5-7B in image-conditioned and text-only accuracy, but its attention IoU and attention recall both decrease. These paired results suggest that pathology tuning can make models better answerers without making them better visual grounders. The complete attention-grounding results on PathVG are provided in Appendix \ref{appA3}.

This finding changes how pathology VLM improvements should be interpreted. If domain training truly strengthened visual-semantic binding, higher answer accuracy would be expected to coincide with larger multimodal gain and better localization of pathological evidence. Instead, pathology-tuned models often improve their ability to predict plausible answers while showing weaker or stagnant grounding metrics. Thus, the central risk is not that pathology training fails to improve accuracy, but that accuracy gains may overstate the real multimodal gains brought by training.

\begin{figure*}[t!]
  \includegraphics[width=1.\textwidth]{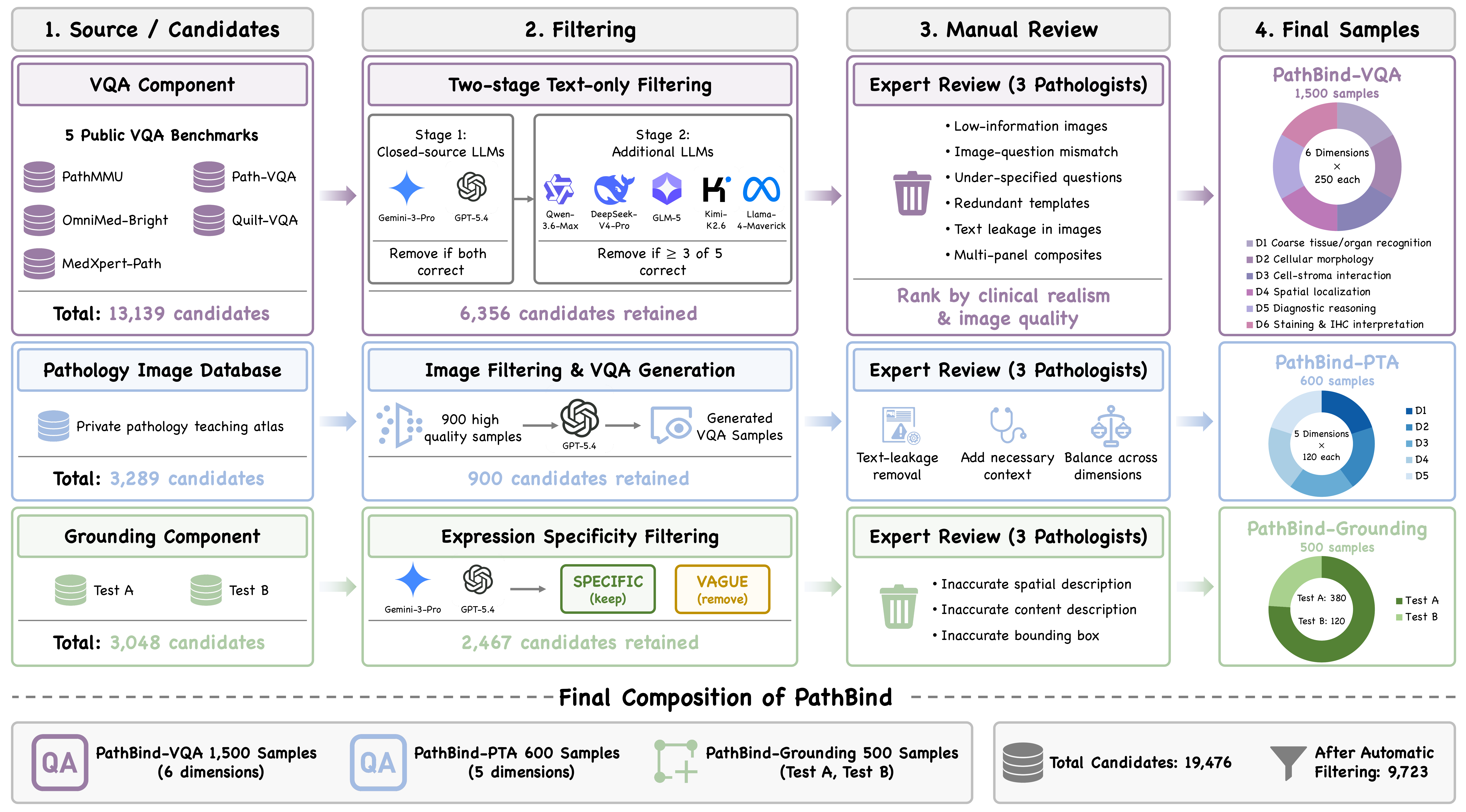}
  \caption{Overview of the PathBind data curation process. PathBind integrates three components: a filtered VQA set, a private pathology VQA set, and a region-level grounding set. Each component undergoes automated filtering followed by expert review to remove non-visual, ambiguous, or low-quality samples.}
  \label{fig3}
\end{figure*}

\subsection{Third Issue: Entity-Level Attention Is Diffuse and Weakly Query-Specific}
\label{3.3}
Behavioral comparisons reveal whether models depend on images, but they do not fully explain whether models bind pathological concepts to visual regions. We therefore examine entity-token attention on PathVG, where pathological expressions are paired with region annotations. A well-grounded model should concentrate attention on the target entity and adjust its attention when the queried entity changes. We measure attention diffuseness using the effective attention support ratio $N_{\mathrm{eff}}/N$, where $N_{\mathrm{eff}}=\exp(H(p))$ is the perplexity of the attention distribution over image patches. Higher values indicate more diffuse attention, and a value close to $1$ corresponds to near-uniform attention.

As shown in Figure \ref{fig2} (b), entity-token attention is highly diffuse across models: the entropy-equivalent attention support exceeds 90\% of the available image patches for all evaluated VLMs, indicating attention distributions close to uniform. This suggests weak spatial concentration when the models process pathological entity queries. Figure \ref{fig2} (c) further shows that attention maps are weakly query-specific. For each image, we compare the attention maps induced by its ground-truth entity query and two alternative entity queries sampled from other examples, and report their mean pairwise correlation. Cross-query attention correlations remain high for most models, suggesting that different pathological entity queries often induce similar attention patterns. Taken together, these results reveal a substantial grounding gap: pathology VLMs may touch relevant regions, but their visual evidence remains broad, loose, and insufficiently entity-specific. Detailed attention extraction procedures, region-overlap metrics, effective attention support, and cross-query attention correlation are described in Appendix \ref{appC}.


\begin{table*}[t!]
\renewcommand{\arraystretch}{1.2}
\renewcommand{\tabcolsep}{5pt}
\centering
\resizebox{\textwidth}{!}{
\begin{tabular}{l|l|cccccc|ccccc|c}
\toprule[1pt]
\multirow{2}{*}{Model} & \multirow{2}{*}{Strategy} & \multicolumn{6}{c|}{PathBind-VQA (1{,}500)} & \multicolumn{5}{c|}{PathBind-PTA (600)} & \multirow{2}{*}{Avg.} \\ \cline{3-13}
 & & D1 & D2 & D3 & D4 & D5 & D6 & D1 & D2 & D3 & D4 & D5 & \\ \midrule[0.7pt]
\multicolumn{14}{c}{\textit{Baseline}} \\ \midrule[0.7pt]
Random Choice & - & 25.0 & 25.0 & 25.0 & 25.0 & 25.0 & 25.0 & 25.0 & 25.0 & 25.0 & 25.0 & 25.0 & 25.0 \\ \midrule[0.7pt]
\multicolumn{14}{c}{\textit{Closed-source VLMs}} \\ \midrule[0.7pt]
\multirow{2}{*}{GPT-5.4} & VLM & 73.2\textcolor{black}{$_{\uparrow 23.6}$} & \textbf{75.2}\textcolor{black}{$_{\uparrow 27.6}$} & \textbf{76.8}\textcolor{black}{$_{\uparrow 24.0}$} & 69.2\textcolor{black}{$_{\uparrow 16.8}$} & \textbf{70.0}\textcolor{black}{$_{\uparrow 24.4}$} & 71.2\textcolor{black}{$_{\uparrow 20.4}$} & 53.3\textcolor{black}{$_{\uparrow 21.6}$} & \textbf{83.3}\textcolor{black}{$_{\uparrow 19.1}$} & \textbf{87.5}\textcolor{black}{$_{\uparrow 24.2}$} & \textbf{85.0}\textcolor{black}{$_{\uparrow 24.2}$} & 70.0\textcolor{black}{$_{\uparrow 39.2}$} & 74.1\textcolor{black}{$_{\uparrow 24.1}$} \\
 & VLM-text & 49.6 & 47.6 & 52.8 & 52.4 & 45.6 & 50.8 & 31.7 & 64.2 & 63.3 & 60.8 & 30.8 & 50.0 \\
\multirow{2}{*}{Gemini-3-Pro} & VLM & \textbf{80.0}\textcolor{black}{$_{\uparrow 21.2}$} & 74.8\textcolor{black}{$_{\uparrow 19.2}$} & 71.6\textcolor{black}{$_{\uparrow 17.6}$} & \textbf{75.2}\textcolor{black}{$_{\uparrow 20.8}$} & 67.2\textcolor{black}{$_{\uparrow 16.0}$} & \textbf{72.0}\textcolor{black}{$_{\uparrow 15.2}$} & \textbf{68.3}\textcolor{black}{$_{\uparrow 20.0}$} & 82.5\textcolor{black}{$_{\uparrow 29.2}$} & 85.8\textcolor{black}{$_{\uparrow 21.6}$} & \textbf{85.0}\textcolor{black}{$_{\uparrow 21.7}$} & \textbf{79.2}\textcolor{black}{$_{\uparrow 42.5}$} & \textbf{76.5}\textcolor{black}{$_{\uparrow 22.3}$} \\
 & VLM-text & \underline{58.8} & \underline{55.6} & \underline{54.0} & \underline{54.4} & \underline{51.2} & \underline{56.8} & \underline{48.3} & 53.3 & 64.2 & 63.3 & 36.7 & \underline{54.2} \\ \midrule[0.7pt]
\multicolumn{14}{c}{\textit{Open-source general VLMs and pathology VLMs}} \\ \midrule[0.7pt]
\multirow{2}{*}{LLaVA-1.5-7B} & VLM & 40.4\textcolor{black}{$_{\uparrow 5.2}$} & 40.8\textcolor{black}{$_{\uparrow 4.8}$} & 48.8\textcolor{black}{$_{\uparrow 8.8}$} & 41.2\textcolor{black}{$_{\uparrow 9.6}$} & 35.2\textcolor{black}{$_{\uparrow 8.8}$} & 34.0\textcolor{black}{$_{\uparrow 12.4}$} & 38.3\textcolor{black}{$_{\uparrow 4.1}$} & 60.0\textcolor{black}{$_{\uparrow 8.3}$} & 67.5\textcolor{black}{$_{\uparrow 28.3}$} & 71.7\textcolor{black}{$_{\uparrow 13.4}$} & 29.2\textcolor{black}{$_{\uparrow 6.7}$} & 46.1\textcolor{black}{$_{\uparrow 10.0}$} \\
 & VLM-text & 35.2 & 36.0 & 40.0 & 31.6 & 26.4 & 21.6 & 34.2 & 51.7 & 39.2 & 58.3 & 22.5 & 36.1 \\
\multirow{2}{*}{Quilt-LLaVA} & VLM & 46.4\textcolor{black}{$_{\uparrow 8.8}$} & 49.6\textcolor{black}{$_{\uparrow 6.4}$} & 42.4\textcolor{black}{$_{\uparrow 4.0}$} & 49.2\textcolor{black}{$_{\uparrow 9.2}$} & 34.0\textcolor{black}{$_{\uparrow 7.2}$} & 34.8\textcolor{black}{$_{\uparrow 10.0}$} & 32.5\textcolor{black}{$_{\uparrow 1.7}$} & 57.5\textcolor{black}{$_{\uparrow 9.2}$} & 70.8\textcolor{black}{$_{\uparrow 21.6}$} & 71.7\textcolor{black}{$_{\uparrow 14.2}$} & 34.2\textcolor{black}{$_{\uparrow 3.4}$} & 47.6\textcolor{black}{$_{\uparrow 8.7}$} \\
 & VLM-text & 37.6 & 43.2 & 38.4 & 40.0 & 26.8 & 24.8 & 30.8 & 48.3 & 49.2 & 57.5 & 30.8 & 38.9 \\
\multirow{2}{*}{LLaVA-1.5-13B} & VLM & 42.0\textcolor{black}{$_{\uparrow 16.4}$} & 45.6\textcolor{black}{$_{\uparrow 11.2}$} & 44.0\textcolor{black}{$_{\uparrow 10.0}$} & 42.0\textcolor{black}{$_{\uparrow 14.4}$} & 37.6\textcolor{black}{$_{\uparrow 11.2}$} & 37.6\textcolor{black}{$_{\uparrow 11.2}$} & 48.3\textcolor{black}{$_{\uparrow 10.8}$} & 63.3\textcolor{black}{$_{\uparrow 10.0}$} & 71.7\textcolor{black}{$_{\uparrow 18.4}$} & 74.2\textcolor{black}{$_{\uparrow 15.0}$} & 35.8\textcolor{black}{$_{\uparrow 11.6}$} & 49.3\textcolor{black}{$_{\uparrow 12.8}$} \\
 & VLM-text & 25.6 & 34.4 & 34.0 & 27.6 & 26.4 & 26.4 & 37.5 & 53.3 & 53.3 & 59.2 & 24.2 & 36.5 \\
\multirow{2}{*}{PathGen-LLaVA} & VLM & 55.6\textcolor{black}{$_{\uparrow 18.8}$} & 53.2\textcolor{black}{$_{\uparrow 15.6}$} & 56.0\textcolor{black}{$_{\uparrow 13.6}$} & 58.8\textcolor{black}{$_{\uparrow 16.4}$} & 54.8\textcolor{black}{$_{\uparrow 15.2}$} & 60.8\textcolor{black}{$_{\uparrow 17.2}$} & 55.8\textcolor{black}{$_{\uparrow 19.1}$} & 80.0\textcolor{black}{$_{\uparrow 13.3}$} & 80.0\textcolor{black}{$_{\uparrow 17.5}$} & 78.3\textcolor{black}{$_{\uparrow 16.6}$} & 57.5\textcolor{black}{$_{\uparrow 22.5}$} & 62.8\textcolor{black}{$_{\uparrow 16.9}$} \\
 & VLM-text & 36.8 & 37.6 & 42.4 & 42.4 & 39.6 & 43.6 & 36.7 & \underline{66.7} & 62.5 & 61.7 & 35.0 & 45.9 \\
\multirow{2}{*}{Qwen2-VL-7B} & VLM & 40.0\textcolor{black}{$_{\uparrow 14.8}$} & 38.4\textcolor{black}{$_{\uparrow 13.6}$} & 40.4\textcolor{black}{$_{\uparrow 15.6}$} & 41.2\textcolor{black}{$_{\uparrow 17.2}$} & 30.8\textcolor{black}{$_{\uparrow 10.0}$} & 40.4\textcolor{black}{$_{\uparrow 18.8}$} & 51.7\textcolor{black}{$_{\uparrow 16.7}$} & 75.8\textcolor{black}{$_{\uparrow 10.0}$} & 63.3\textcolor{black}{$_{\uparrow 12.5}$} & 77.5\textcolor{black}{$_{\uparrow 22.5}$} & 47.5\textcolor{black}{$_{\uparrow 15.0}$} & 49.7\textcolor{black}{$_{\uparrow 15.1}$} \\
 & VLM-text & 25.2 & 24.8 & 24.8 & 24.0 & 20.8 & 21.6 & 35.0 & 65.8 & 50.8 & 55.0 & 32.5 & 34.6 \\
\multirow{2}{*}{HuatuoGPT-Vision-7B} & VLM & 46.4\textcolor{black}{$_{\uparrow 16.0}$} & 49.6\textcolor{black}{$_{\uparrow 23.2}$} & 48.8\textcolor{black}{$_{\uparrow 20.0}$} & 46.8\textcolor{black}{$_{\uparrow 17.6}$} & 31.6\textcolor{black}{$_{\uparrow 10.4}$} & 40.8\textcolor{black}{$_{\uparrow 10.4}$} & 49.2\textcolor{black}{$_{\uparrow 17.5}$} & 69.2\textcolor{black}{$_{\uparrow 17.5}$} & 75.0\textcolor{black}{$_{\uparrow 26.7}$} & 70.0\textcolor{black}{$_{\uparrow 15.8}$} & 52.5\textcolor{black}{$_{\uparrow 23.3}$} & 52.7\textcolor{black}{$_{\uparrow 18.0}$} \\
 & VLM-text & 30.4 & 26.4 & 28.8 & 29.2 & 21.2 & 30.4 & 31.7 & 51.7 & 48.3 & 54.2 & 29.2 & 34.7 \\
\multirow{2}{*}{Qwen2.5-VL-7B} & VLM & 44.0\textcolor{black}{$_{\uparrow 22.8}$} & 43.2\textcolor{black}{$_{\uparrow 14.8}$} & 42.4\textcolor{black}{$_{\uparrow 16.4}$} & 44.4\textcolor{black}{$_{\uparrow 25.2}$} & 42.4\textcolor{black}{$_{\uparrow 20.8}$} & 44.0\textcolor{black}{$_{\uparrow 24.8}$} & 56.7\textcolor{black}{$_{\uparrow 20.9}$} & 75.0\textcolor{black}{$_{\uparrow 12.5}$} & 75.8\textcolor{black}{$_{\uparrow 10.8}$} & 81.7\textcolor{black}{$_{\uparrow 18.4}$} & 41.7\textcolor{black}{$_{\uparrow 16.7}$} & 53.8\textcolor{black}{$_{\uparrow 18.6}$} \\
 & VLM-text & 21.2 & 28.4 & 26.0 & 19.2 & 21.6 & 19.2 & 35.8 & 62.5 & 65.0 & 63.3 & 25.0 & 35.2 \\
\multirow{2}{*}{Patho-R1-7B} & VLM & 60.4\textcolor{black}{$_{\uparrow 21.6}$} & 52.8\textcolor{black}{$_{\uparrow 17.6}$} & 52.8\textcolor{black}{$_{\uparrow 18.8}$} & 55.2\textcolor{black}{$_{\uparrow 19.6}$} & 60.0\textcolor{black}{$_{\uparrow 21.6}$} & 59.6\textcolor{black}{$_{\uparrow 23.6}$} & 62.5\textcolor{black}{$_{\uparrow 22.5}$} & 78.3\textcolor{black}{$_{\uparrow 15.0}$} & 81.7\textcolor{black}{$_{\uparrow 11.7}$} & 75.0\textcolor{black}{$_{\uparrow 6.7}$} & 60.8\textcolor{black}{$_{\uparrow 19.1}$} & 63.6\textcolor{black}{$_{\uparrow 18.0}$} \\
 & VLM-text & 38.8 & 35.2 & 34.0 & 35.6 & 38.4 & 36.0 & 40.0 & 63.3 & \underline{70.0} & \underline{68.3} & \underline{41.7} & 45.6 \\
\bottomrule[1pt]
\end{tabular}}
\caption{Evaluations of various VLMs on PathBind-VQA and PathBind-PTA across diagnostic dimensions. ``VLM'' denotes evaluating with images, ``VLM-text'' evaluates without images. PathBind-VQA covers six dimensions (D1-D6); PathBind-PTA covers the first five (D1-D5). The highest results of ``VLM'' and ``VLM-text'' per column are highlighted in \textbf{bold} and \underline{underlined}. The numbers with arrows indicate the improvement of “VLM” relative to “VLM-text”.}
\label{table2}
\end{table*}

\section{PathBind}

\subsection{Data Curation Process}

\paragraph{Criteria for data curation.}
PathBind is constructed according to three criteria: (1) strong visual dependence, requiring models to inspect histopathology images rather than infer answers from textual priors alone; (2) entity-level visual-semantic grounding, requiring pathological concepts to correspond to identifiable microscopic evidence; and (3) broad capability coverage across complementary pathology reasoning skills.

\paragraph{Data filter.}

As shown in Figure \ref{fig3}, PathBind integrates public pathology VQA benchmarks, PathVG, and a private pathology teaching atlas. For PathBind-VQA, we collect 13,139 questions from PathMMU, Path-VQA, OmniMed-Bright, Quilt-VQA, and MedXpert-Path, remove invalid or duplicated samples, and apply two-stage text-only filtering. Samples answered correctly by both closed-source LLMs \cite{GPT-5.4, Gemini3pro} are removed in the first stage; the remaining samples are further filtered by five additional LLMs \cite{qwen36_max_preview, glm5team2026glm5, moonshotai2026kimik26, deepseekai2026deepseekv4pro, meta2025llama4}, removing those answered correctly by at least three models. This yields 6,356 candidates. For PathBind-Grounding, we begin with 3,048 PathVG samples and retain 2,467 expressions that describe a specific, localizable pathological entity or visual pattern. For PathBind-PTA, high-quality images are selected from the teaching atlas, and GPT-5.4 assists in generating 900 multiple-choice candidates. Further filtering details are provided in Appendix \ref{appB1}-\ref{appB3}.

\paragraph{Manual review.}
After automated filtering, all candidates were independently reviewed by two attending pathologists. Disagreements were resolved through joint discussion with a senior consultant pathologist, and a final consensus decision was reached. For PathBind-VQA, the reviewers excluded samples with low-information images, image-question mismatch, under-specified questions, redundant templates, text leakage within images, or multi-panel composite images. The retained samples were ranked by clinical realism and image quality, yielding \textbf{1,500} questions balanced across six diagnostic dimensions. For PathBind-Grounding, the reviewers assessed the retained PathVG candidates for expression specificity, expression-bounding-box consistency, and bounding-box accuracy. Samples with inaccurate spatial descriptions, inaccurate content descriptions, or inaccurate bounding boxes were excluded, yielding \textbf{500} samples, including 380 from testA and 120 from testB. For PathBind-PTA, the reviewers assessed the generated candidates, removed ambiguous cases and samples with text leakage, balanced the diagnostic dimensions and answer options, and added necessary anatomical or system context when the diagnosis would otherwise be under-specified, resulting in \textbf{600} questions across five dimensions. Together, these components form \textbf{PathBind}, a diagnostic benchmark comprising \textbf{2,600} samples for evaluating visual-semantic binding in pathology VLMs. Detailed reviewer qualifications, agreement statistics, and consensus procedures are provided in Appendix~\ref{appB5}.

\subsection{Diagnostic Dimensions}
We define six diagnostic dimensions for evaluating pathology VLMs across different levels of visual reasoning: coarse tissue or organ recognition, cellular morphology, cell-stroma interaction, spatial localization, diagnostic reasoning, and staining or immunohistochemistry (IHC) interpretation. As shown in Figure~\ref{fig3}, PathBind-VQA contains 1,500 samples evenly distributed across the six dimensions, while PathBind-PTA contains 600 samples across the first five dimensions after excluding staining and IHC. Detailed definitions of each diagnostic dimension are provided in Appendix~\ref{appB4}.

\subsection{Evaluation Protocol}
For VQA evaluation, PathBind reports both image-conditioned and text-only accuracy. The improvement from text-only to image-conditioned inference is shown directly in the tables and noted as multimodal gain to indicate the benefits from visual evidence. For grounding evaluation, PathBind reports IoU, precision, and recall between model-induced grounding regions and annotated pathological regions, with detailed definitions provided in Appendix~\ref{appC2}.

\section{Experiments}
In this section, we conduct a systematic evaluation of PathBind. Across the full study, we evaluate 18 closed-source, general-purpose, and pathology-tuned VLMs. The main PathBind evaluation reports results for 10 representative models. Full results and experimental details are reported in Appendices \ref{appA}-\ref{appE}.

\subsection{Results Analysis of PathBind}
We present the VQA results on PathBind-VQA and PathBind-PTA in Table \ref{table2}, and report the grounding results on PathBind-Grounding in Table \ref{table3}. We summarize the principal observations below.

\begin{table}[t!]
\renewcommand{\arraystretch}{1.2}
\renewcommand{\tabcolsep}{8pt}
\centering
\resizebox{\columnwidth}{!}{
\begin{tabular}{l|cccccc}
\toprule[1pt]
\multirow{2}{*}{Model} & \multicolumn{3}{c}{Test A (380)} & \multicolumn{3}{c}{Test B (120)} \\ \cline{2-7}
 & IoU & Pre & Rec & IoU & Pre & Rec \\ \midrule[0.7pt]
InternVL3-8B & 10.0 & 13.3 & 32.4 & 3.9 & 4.0 & 40.5 \\
Qwen3-VL-8B & 21.6 & 27.2 & 68.6 & \textbf{9.9} & \textbf{10.1} & \textbf{80.3} \\
LLaVA-1.5-7B & 17.6 & 23.7 & 51.6 & 3.0 & 3.1 & 26.7 \\
Quilt-LLaVA & 12.4 & 18.3 & 35.7 & 4.0 & 4.2 & 32.3 \\
LLaVA-1.5-13B & 16.4 & 22.2 & 49.6 & 6.2 & 6.4 & 48.8 \\
PathGen-LLaVA & 17.5 & 24.3 & 53.9 & 7.9 & 8.0 & 56.3 \\
Qwen2-VL-7B & 14.9 & 19.7 & 48.8 & 5.4 & 5.5 & 57.2 \\
HuatuoGPT-Vision-7B & 16.8 & 22.4 & 51.9 & 5.9 & 6.1 & 55.3 \\
Qwen2.5-VL-7B & \textbf{23.3} & \textbf{29.4} & \textbf{71.6} & 8.2 & 8.4 & 77.9 \\
Patho-R1-7B & 17.4 & 23.1 & 55.9 & 7.2 & 7.5 & 63.3 \\
\bottomrule[1pt]
\end{tabular}}
\caption{Attention-grounding results on PathBind-Grounding. We report the IoU, Precision, and Recall between entity-token attention maps and ground-truth bounding boxes for pathological expressions on the 380 testA and 120 testB samples that passed expert review. The best results are highlighted in \textbf{bold}.}
\label{table3}
\end{table}

\begin{figure*}[t!]
  \includegraphics[width=1.\textwidth]{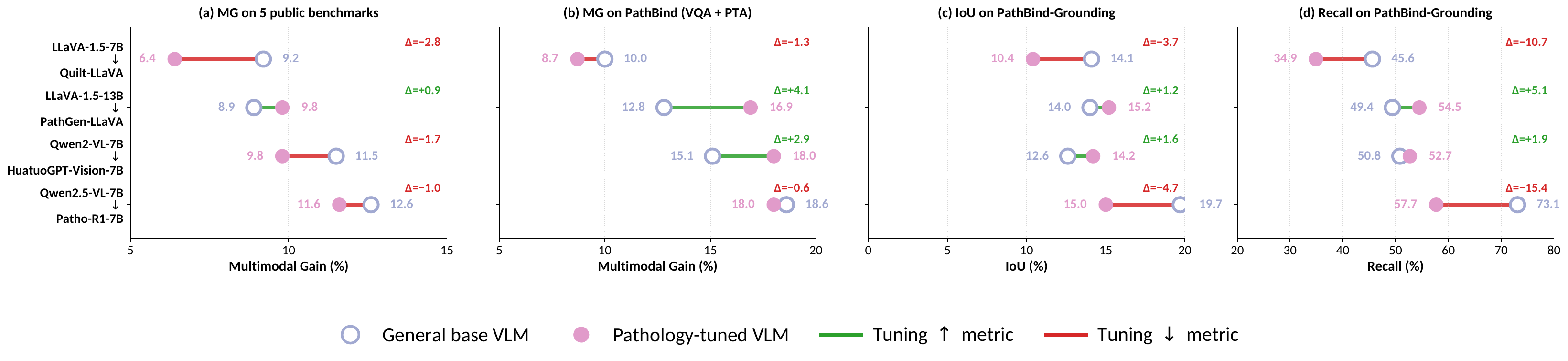}
  \caption{General-VLM $\to$ pathology-tuned pairs across four metrics. For each pair, we compare (a) Multimodal Gain on five raw public benchmarks; (b) MG on PathBind (VQA + PTA); (c) IoU and (d) Recall on PathBind-Grounding. Green/red lines mark pairs where the tuned model scores above/below its general model.}
  \label{fig4}
\end{figure*}

\paragraph{Observation from VQA performance.}
The closed-source VLMs achieve the strongest answer-side performance. Gemini-3-Pro obtains the highest average accuracy of 76.5\%, followed by GPT-5.4 with 74.1\%. Among the open-source models, Patho-R1 and PathGen-LLaVA achieve the best average accuracies of 63.6\% and 62.8\%, respectively, substantially outperforming the remaining open-source models. 

Performance also varies across diagnostic dimensions. Models that perform well on coarse tissue recognition do not necessarily maintain the same advantage on cellular morphology, spatial localization, or diagnostic reasoning. This variation supports the need for dimension-level reporting rather than relying solely on one aggregate score. PathBind-PTA generally yields higher image-conditioned accuracy than PathBind-VQA, particularly for cellular morphology and spatial localization, while several models remain weak on diagnostic reasoning. This suggests that controlled teaching-atlas images facilitate visual recognition but do not eliminate the difficulty of integrating visual evidence into diagnostic decisions.

\paragraph{Observation from visual dependence.}
Although PathBind removes samples that can be consistently solved by multiple VLMs without images, models still retain non-trivial text-only performance. Gemini-3-Pro and GPT-5.4 achieve average text-only accuracies of 54.2\% and 50.0\%, respectively, while the strongest pathology-tuned open-source models also exceed 45\%. This is expected because pathology terminology and answer choices can still provide partial semantic cues even when a question cannot be reliably solved without its image. More importantly, images provide substantially larger gains on PathBind than on the original public benchmarks. As shown in Figure \ref{fig4}, all eight models obtain greater multimodal gains on PathBind than on the five raw public benchmarks. For example, PathGen-LLaVA improves from a 9.8 multimodal gain on the existing benchmarks to 16.9 percentage points on PathBind, while HuatuoGPT-Vision-7B improves from 9.8 to 18.0 percentage points. These results indicate that the filtering and expert-review process increases visual dependence, even though pathology language priors cannot be eliminated.

\paragraph{Observation from grounding performance.}
The grounding results reveal a markedly different model ranking from the VQA results. On testA, Qwen2.5-VL-7B achieves the highest IoU of 23.3\% and precision of 29.4\%, while Qwen3-VL-8B achieves the strongest results on testB, with 9.9\% IoU and 80.3\% recall. Notably, general-purpose VLMs remain competitive with, and often outperform, pathology-tuned models in region-level grounding. Across all models, IoU and precision remain low despite relatively high recall. For example, Qwen3-VL-8B achieves 80.3\% recall on testB, but its IoU and precision are only 9.9\% and 10.1\%, respectively. This discrepancy indicates that attention maps frequently touch the annotated pathological region while also spreading over broad irrelevant areas. Consequently, high recall should not be interpreted as precise entity grounding. These results are consistent with the diffuse and weakly query-specific attention patterns identified in Section \ref{3.3}.

\subsection{Analysis of Domain Training Effects}
Pathology-specific training improves image-conditioned accuracy for all four model pairs, but its effect on visual dependence remains inconsistent. As shown in Figure \ref{fig4}, on the five public benchmarks, three tuned models show lower multimodal gain than their general-purpose models, with only PathGen-LLaVA improving slightly by 0.9 points. PathBind yields more positive results: PathGen-LLaVA and HuatuoGPT-Vision improve multimodal gain by 4.1 points and 2.9 points, whereas Quilt-LLaVA and Patho-R1 remain below their general models by 1.3 points and 0.6 points. Thus, pathology training can strengthen image use in some model families, but higher answer accuracy does not necessarily imply a larger contribution from visual evidence.

The grounding comparisons show a similarly mixed pattern. PathGen-LLaVA improves over LLaVA-1.5-13B by 1.2 points in IoU and 5.1 points in recall, while HuatuoGPT-Vision also provides small gains of 1.6 points and 1.9 points over Qwen2-VL-7B. In contrast, Quilt-LLaVA decreases by 3.7 points in IoU and 10.7 points in recall, and Patho-R1 shows the largest regression, with IoU and recall falling by 4.7 points and 15.4 points relative to Qwen2.5-VL-7B. Overall, pathology-specific training improves answer prediction more consistently than visual dependence. These results support the domain training illusion: answer-side improvements can overstate the multimodal gains introduced by domain-specific training.

\section{Conclusion}

We revisit pathology VLM evaluation by distinguishing answer correctness from visual dependence and entity-level grounding. Our analysis reveals three issues with current evaluation: substantial text-only solvability, inconsistent multimodal gains from pathology-specific training, and weakly query-specific entity attention. To address this, we introduce PathBind, a diagnostic benchmark that combines filtered public VQA questions, teaching-atlas VQA, and expert-curated region-level grounding samples. Experiments show that strong answer performance does not consistently translate into precise visual grounding, while pathology-tuned models do not uniformly outperform their general-purpose models in visual dependence. These findings demonstrate that answer accuracy alone is insufficient for assessing pathology understanding and motivate evaluation protocols that jointly measure answer correctness, image dependence, and localized visual evidence.

\bibliography{aaai2027}

\clearpage
\setcounter{figure}{0} 
\renewcommand{\thefigure}{\thesection.\arabic{figure}}
\setcounter{table}{0}
\renewcommand{\thetable}{\thesection.\arabic{table}}
\appendix

\section{Additional Experimental Results and Analyses}
\label{appA}
In this section, we provide additional experimental results and analyses that complement the findings in the main paper. Specifically, we report (1) expanded evaluation results on additional VLMs, (2) image ablation experiments for examining visual dependence, and (3) full grounding results on PathVG to support the observations in Section \ref{sec3}.

\subsection{More Results on Pathology VQA Benchmarks and PathBind}
\label{appA1}
\paragraph{Full evaluation results on five pathology VQA benchmarks.}
Table \ref{table A1} extends Table \ref{table1} in the main paper by including additional general-purpose VLMs. Following the same protocol, we evaluate each model in both image-conditioned and text-only settings across multiple pathology VQA benchmarks.

Consistent with the observations in the main paper, many models retain substantial performance without visual input. This phenomenon is observed across both closed-source and open-source models and remains consistent after expanding the evaluated model set. These observations suggest that benchmark accuracy may partially reflect textual shortcuts and domain knowledge rather than genuine visual understanding. Therefore, answer accuracy alone may not provide a faithful estimate of pathology visual-semantic capabilities.

\paragraph{Full evaluation results on PathBind.}
Table~\ref{table A_PathBind} extends the main results in Table \ref{table2} by reporting evaluations of all 18 VLMs on PathBind-VQA and PathBind-PTA across different diagnostic dimensions. Following the same evaluation protocol, we report both image-conditioned and text-only accuracy, together with the corresponding improvement from visual input.

Consistent with the main observations, closed-source VLMs achieve the strongest answer-side performance, while pathology-specific models do not consistently exhibit larger multimodal gains than their general models. The results across diagnostic dimensions further demonstrate that performance varies substantially depending on the required pathology reasoning ability. These findings support the conclusion that high answer accuracy alone does not necessarily indicate stronger visual-semantic binding.

\begin{table*}[t!]
\renewcommand{\arraystretch}{1.2}
\renewcommand{\tabcolsep}{8pt}
\centering
\resizebox{\textwidth}{!}
{\begin{tabular}{l|l|cccccc|c}
\toprule[1pt]
\multirow{2}{*}{Model}               & \multirow{2}{*}{Strategy} & \multicolumn{2}{c}{PathMMU} & \multicolumn{2}{c}{YorN} & MedXpert & OmniMed & \multirow{2}{*}{Avg.} \\ \cline{3-8} 
                                     &                           & Test-tiny     & Test-all    & Quilt-VQA   & Path-VQA   & Path       & Bright     \\ \midrule[0.7pt]
\multicolumn{9}{c}{\textit{Baseline}}                                                                                                                        \\ \midrule[0.7pt]
Random Choice                        & -                         & 23.4          & 24.2        & 50.0        & 50.0       & 21.3       & 24.2   & 32.2    \\ \midrule[0.7pt]
\multicolumn{9}{c}{\textit{Closed-source VLMs}}                                                                                                             \\ \midrule[0.7pt]
\multirow{2}{*}{GPT-5.4}             & VLM                & 70.8\textcolor{black}{$_{\uparrow 25.1}$}              & 68.6\textcolor{black}{$_{\uparrow 21.6}$}           & 68.2\textcolor{black}{$_{\uparrow 25.7}$}            & 71.1\textcolor{black}{$_{\uparrow 23.4}$}           & 68.9\textcolor{black}{$_{\uparrow 14.4}$}           & 61.2\textcolor{black}{$_{\uparrow 17.2}$}   & 68.1\textcolor{black}{$_{\uparrow 21.2}$}        \\
                                     & VLM-text                 & 45.7              & 47.0            & 42.5            & 47.7           & \underline{54.5}           & 44.0   & 46.9       \\
\multirow{2}{*}{Gemini-3-Pro}        & VLM                &\textbf{77.8}\textcolor{black}{$_{\uparrow 26.5}$}            & \textbf{75.6}\textcolor{black}{$_{\uparrow 25.1}$}            & \textbf{73.5}\textcolor{black}{$_{\uparrow 24.8}$}            & \textbf{77.5}\textcolor{black}{$_{\uparrow 17.8}$}           & \textbf{80.0}\textcolor{black}{$_{\uparrow 25.6}$}           & \textbf{71.8}\textcolor{black}{$_{\uparrow 15.7}$}     &  \textbf{76.0}\textcolor{black}{$_{\uparrow 22.5}$}    \\
                                     & VLM-text                 & \underline{51.3}              & \underline{50.5}            & 48.7            & 59.7           & 54.4           & 56.1     & \underline{53.5}     \\ \midrule[0.7pt]
\multicolumn{9}{c}{\textit{Open-source general VLMs and pathology VLMs}}                                                                                     \\ \midrule[0.7pt]
\multirow{2}{*}{LLaVA-1.5-7B}        & VLM                &  37.0\textcolor{black}{$_{\uparrow 5.3}$}             & 36.1\textcolor{black}{$_{\uparrow 5.4}$}            & 56.6\textcolor{black}{$_{\uparrow 14.6}$}            & 49.5\textcolor{black}{$_{\uparrow 9.9}$}           & 22.2\textcolor{black}{$_{\uparrow 7.8}$}           & 46.5\textcolor{black}{$_{\uparrow 12.3}$}      & 41.3\textcolor{black}{$_{\uparrow 9.2}$}    \\
                                     & VLM-text                 & 31.7              & 30.7            & 42.0            & 39.6           & 14.4           & 34.2   & 32.1        \\
\multirow{2}{*}{Quilt-LLaVA}         & VLM                &  42.0\textcolor{black}{$_{\uparrow 6.7}$}             & 42.2\textcolor{black}{$_{\uparrow 5.4}$}            & 64.3\textcolor{black}{$_{\uparrow 10.3}$}            & 57.3\textcolor{black}{$_{\uparrow 1.1}$}           & 21.1\textcolor{black}{$_{\uparrow 4.4}$}           & 49.8\textcolor{black}{$_{\uparrow 10.6}$}        & 46.1\textcolor{black}{$_{\uparrow 6.4}$}    \\
                                     & VLM-text                 & 35.3              & 36.8            & 54.0            & 56.2           & 16.7           & 39.2     & 39.7     \\
\multirow{2}{*}{LLaVA-1.5-13B}       & VLM                &  36.3\textcolor{black}{$_{\uparrow 4.2}$}             & 37.2\textcolor{black}{$_{\uparrow 5.1}$}            &  58.3\textcolor{black}{$_{\uparrow 14.1}$}           & 60.8\textcolor{black}{$_{\uparrow 13.4}$}           & 18.9\textcolor{black}{$_{\uparrow 4.6}$}           & 45.2\textcolor{black}{$_{\uparrow 11.9}$}         & 42.8\textcolor{black}{$_{\uparrow 8.9}$} \\
                                     & VLM-text                 & 32.1              & 32.1            & 44.2            & 47.4           & 14.3           & 33.3     & 33.9     \\
\multirow{2}{*}{PathGen-LLaVA}       & VLM                &  62.6\textcolor{black}{$_{\uparrow 11.4}$}             & 60.4\textcolor{black}{$_{\uparrow 11.3}$}            & 62.3\textcolor{black}{$_{\uparrow 13.7}$}            & 63.1\textcolor{black}{$_{\uparrow 4.9}$}           & 30.0\textcolor{black}{$_{\uparrow 6.7}$}           & 68.5\textcolor{black}{$_{\uparrow 11.2}$}         & 57.8\textcolor{black}{$_{\uparrow 9.8}$} \\
                                     & VLM-text                 & 51.2              & 49.1            & 48.6            & 58.2           & 23.3           & 57.3    & 48.0      \\
\multirow{2}{*}{Qwen2-VL-7B}         & VLM                & 48.6\textcolor{black}{$_{\uparrow 13.5}$}              & 47.0\textcolor{black}{$_{\uparrow 12.3}$}            & 52.6\textcolor{black}{$_{\uparrow 14.3}$}            & 60.6\textcolor{black}{$_{\uparrow 14.1}$}           & 18.9\textcolor{black}{$_{\uparrow 5.6}$}           & 51.9\textcolor{black}{$_{\uparrow 9.0}$}          & 46.6\textcolor{black}{$_{\uparrow 11.5}$} \\
                                     & VLM-text                 & 35.1              & 34.7            & 38.3            & 46.5           & 13.3           & 42.9    & 35.1      \\
\multirow{2}{*}{HuatuoGPT-Vision-7B} & VLM                & 57.5\textcolor{black}{$_{\uparrow 11.8}$}              & 56.1\textcolor{black}{$_{\uparrow 12.0}$}            & 62.6\textcolor{black}{$_{\uparrow 15.4}$}            & 60.2\textcolor{black}{$_{\uparrow 10.7}$}           & 20.0\textcolor{black}{$_{\uparrow 2.2}$}           & 44.9\textcolor{black}{$_{\uparrow 7.1}$}         & 50.2\textcolor{black}{$_{\uparrow 9.8}$} \\
                                     & VLM-text                 & 45.7              & 44.1            & 47.2            & 49.5           & 17.8           & 37.8     & 40.4     \\
\multirow{2}{*}{Qwen2.5-VL-7B}       & VLM                & 51.6\textcolor{black}{$_{\uparrow 21.1}$}              & 48.7\textcolor{black}{$_{\uparrow 17.7}$}            &  54.2\textcolor{black}{$_{\uparrow 14.1}$}           & 59.5\textcolor{black}{$_{\uparrow 11.7}$}           & 13.3\textcolor{black}{$_{\uparrow 1.1}$}           & 54.6\textcolor{black}{$_{\uparrow 9.9}$}         & 47.0\textcolor{black}{$_{\uparrow 12.6}$} \\
                                     & VLM-text                 & 30.5              & 31.0            & 40.1            & 47.8           & 12.2           & 44.7   & 34.4       \\
\multirow{2}{*}{Patho-R1-7B}         & VLM                &  63.1\textcolor{black}{$_{\uparrow 15.3}$}             & 62.9\textcolor{black}{$_{\uparrow 18.3}$}            &  64.4\textcolor{black}{$_{\uparrow 11.6}$}           & 55.3\textcolor{black}{$_{\uparrow 12.3}$}           & 16.7\textcolor{black}{$_{\uparrow 0.9}$}           & 69.9\textcolor{black}{$_{\uparrow 11.3}$}        & 55.4\textcolor{black}{$_{\uparrow 11.6}$}   \\
                                     & VLM-text                 & 47.8              & 44.6            & 52.8            & 43.0           & 15.8           & \underline{58.6}    & 43.8      \\
\multirow{2}{*}{Lingshu-7B}         & VLM                &  59.7\textcolor{black}{$_{\uparrow 16.0}$}             & 58.5\textcolor{black}{$_{\uparrow 17.9}$}            &  62.3\textcolor{black}{$_{\uparrow 22.1}$}           & 63.6\textcolor{black}{$_{\uparrow 11.7}$}           & 30.0\textcolor{black}{$_{\uparrow 10.4}$}           & 52.6\textcolor{black}{$_{\uparrow 8.8}$}        & 54.5\textcolor{black}{$_{\uparrow 14.5}$}   \\
                                     & VLM-text                 & 43.7              & 40.6            & 40.2            & 51.9           & 19.6           & 43.8    & 40.0      \\                                      
                                     \midrule[0.7pt]
\multicolumn{9}{c}{\textit{More open-source general VLMs}}                                                                                                             \\ \midrule[0.7pt]
\multirow{2}{*}{Qwen3-VL-8B}             & VLM                & 54.3\textcolor{black}{$_{\uparrow 23.7}$}              & 53.3\textcolor{black}{$_{\uparrow 19.6}$}           & 60.6\textcolor{black}{$_{\uparrow 18.5}$}            & 62.7\textcolor{black}{$_{\uparrow 13.2}$}           & 26.7\textcolor{black}{$_{\uparrow 6.7}$}           & 54.5\textcolor{black}{$_{\uparrow 3.9}$}   & 52.0\textcolor{black}{$_{\uparrow 14.2}$}        \\
                                     & VLM-text                 & 30.6              & 33.7            & 42.1            & 49.5           & 20.0           & 50.6   & 37.8       \\
\multirow{2}{*}{InternVL3-8B}             & VLM                & 51.4\textcolor{black}{$_{\uparrow 18.4}$}              & 52.5\textcolor{black}{$_{\uparrow 18.1}$}           & 60.3\textcolor{black}{$_{\uparrow 18.7}$}            & 61.4\textcolor{black}{$_{\uparrow 10.5}$}           & 24.4\textcolor{black}{$_{\uparrow 3.3}$}           & 60.3\textcolor{black}{$_{\uparrow 10.6}$}   & 51.7\textcolor{black}{$_{\uparrow 13.2}$}        \\
                                     & VLM-text                 & 33.0              & 34.4            & 41.6            & 50.9           & 21.1           & 49.7   & 38.5       \\
\multirow{2}{*}{Llama-3.2-11B}             & VLM                & 49.5\textcolor{black}{$_{\uparrow 17.7}$}              & 48.4\textcolor{black}{$_{\uparrow 16.0}$}           & 53.4\textcolor{black}{$_{\uparrow 16.0}$}            & 60.5\textcolor{black}{$_{\uparrow 11.7}$}           & 16.7\textcolor{black}{$_{\uparrow 3.3}$}           & 54.5\textcolor{black}{$_{\uparrow 1.7}$}   & 47.2\textcolor{black}{$_{\uparrow 11.1}$}        \\
                                     & VLM-text                 & 31.8              & 32.4            & 37.4            & 48.8           & 13.4           & 52.8   & 36.1       \\                 
\multirow{2}{*}{Gemma-3-12B}             & VLM                & 51.2\textcolor{black}{$_{\uparrow 18.4}$}              & 49.4\textcolor{black}{$_{\uparrow 16.8}$}           & 64.1\textcolor{black}{$_{\uparrow 10.8}$}            & 63.4\textcolor{black}{$_{\uparrow 1.2}$}           & 16.7\textcolor{black}{$_{\uparrow 2.2}$}           & 47.3\textcolor{black}{$_{\uparrow 1.8}$}   & 48.7\textcolor{black}{$_{\uparrow 8.5}$}        \\
                                     & VLM-text                 & 32.8              & 32.6            & 53.3            & 62.2           & 14.5           & 45.5   & 40.2       \\                 
\multirow{2}{*}{Gemma-3-27B}             & VLM                & 56.4\textcolor{black}{$_{\uparrow 15.1}$}              & 55.6\textcolor{black}{$_{\uparrow 21.2}$}           & 69.4\textcolor{black}{$_{\uparrow 8.7}$}            & 65.0\textcolor{black}{$_{\uparrow 1.8}$}           & 27.8\textcolor{black}{$_{\uparrow 4.4}$}           & 46.7\textcolor{black}{$_{\uparrow 1.9}$}   & 53.5\textcolor{black}{$_{\uparrow 8.9}$}        \\
                                     & VLM-text                 & 41.3              & 34.4            & \underline{60.7}            & \underline{63.2}           & 23.4           & 44.8   & 44.6       \\                 
\multirow{2}{*}{Qwen3-VL-32B}             & VLM                & 62.6\textcolor{black}{$_{\uparrow 20.7}$}              & 60.4\textcolor{black}{$_{\uparrow 20.1}$}           & 58.6\textcolor{black}{$_{\uparrow 21.0}$}            & 68.8\textcolor{black}{$_{\uparrow 18.9}$}           & 28.7\textcolor{black}{$_{\uparrow 19.5}$}           & 58.7\textcolor{black}{$_{\uparrow 9.4}$}   & 56.3\textcolor{black}{$_{\uparrow 18.3}$}        \\
                                     & VLM-text                 & 41.9              & 40.3            & 37.6            & 49.9           & 9.2           & 49.3   & 38.0       \\    
\multirow{2}{*}{InternVL3-38B}             & VLM                & 55.9\textcolor{black}{$_{\uparrow 19.2}$}              & 54.8\textcolor{black}{$_{\uparrow 17.7}$}           & 63.0\textcolor{black}{$_{\uparrow 23.0}$}            & 71.8\textcolor{black}{$_{\uparrow 18.9}$}           & 24.7\textcolor{black}{$_{\uparrow 4.3}$}           & 48.1\textcolor{black}{$_{\uparrow 10.4}$}   & 53.1\textcolor{black}{$_{\uparrow 15.6}$}        \\
                                     & VLM-text                 & 36.7              & 37.1            & 40.0            & 52.9           & 20.4           & 37.7   & 37.5       \\                                     
                                     \bottomrule[1pt]
\end{tabular}
}
\caption{Full evaluations of 18 VLMs on five popular pathology VQA benchmarks. For the "strategy" column, "VLM" denotes evaluating with images, while "VLM-text" means evaluating without images. We report the results for 2 closed-source VLMs, 5 pathology-tuned VLMs, along with their base models, and 7 open-source general VLMs. The highest results of the "VLM" and "VLM-text" settings across models are highlighted in \textbf{bold} and \underline{underlined}. The numbers with arrows indicate the improvement of “VLM” relative to “VLM-text”.}
\label{table A1}
\end{table*}

\begin{table*}[t!]
\renewcommand{\arraystretch}{1.2}
\renewcommand{\tabcolsep}{5pt}
\centering
\resizebox{\textwidth}{!}{
\begin{tabular}{l|l|cccccc|ccccc|c}
\toprule[1pt]
\multirow{2}{*}{Model} & \multirow{2}{*}{Strategy} & \multicolumn{6}{c|}{PathBind-VQA (1{,}500)} & \multicolumn{5}{c|}{PathBind-PTA (600)} & \multirow{2}{*}{Avg.} \\ \cline{3-13}
 & & D1 & D2 & D3 & D4 & D5 & D6 & D1 & D2 & D3 & D4 & D5 & \\ \midrule[0.7pt]
\multicolumn{14}{c}{\textit{Baseline}} \\ \midrule[0.7pt]
Random Choice & - & 25.0 & 25.0 & 25.0 & 25.0 & 25.0 & 25.0 & 25.0 & 25.0 & 25.0 & 25.0 & 25.0 & 25.0 \\ \midrule[0.7pt]
\multicolumn{14}{c}{\textit{Closed-source VLMs}} \\ \midrule[0.7pt]
\multirow{2}{*}{GPT-5.4} & VLM & 73.2\textcolor{black}{$_{\uparrow 23.6}$} & \textbf{75.2}\textcolor{black}{$_{\uparrow 27.6}$} & \textbf{76.8}\textcolor{black}{$_{\uparrow 24.0}$} & 69.2\textcolor{black}{$_{\uparrow 16.8}$} & \textbf{70.0}\textcolor{black}{$_{\uparrow 24.4}$} & 71.2\textcolor{black}{$_{\uparrow 20.4}$} & 53.3\textcolor{black}{$_{\uparrow 21.6}$} & 83.3\textcolor{black}{$_{\uparrow 19.1}$} & \textbf{87.5}\textcolor{black}{$_{\uparrow 24.2}$} & \textbf{85.0}\textcolor{black}{$_{\uparrow 24.2}$} & 70.0\textcolor{black}{$_{\uparrow 39.2}$} & 74.1\textcolor{black}{$_{\uparrow 24.1}$} \\
 & VLM-text & 49.6 & 47.6 & 52.8 & 52.4 & 45.6 & 50.8 & 31.7 & 64.2 & 63.3 & 60.8 & 30.8 & 50.0 \\
\multirow{2}{*}{Gemini-3-Pro} & VLM & \textbf{80.0}\textcolor{black}{$_{\uparrow 21.2}$} & 74.8\textcolor{black}{$_{\uparrow 19.2}$} & 71.6\textcolor{black}{$_{\uparrow 17.6}$} & \textbf{75.2}\textcolor{black}{$_{\uparrow 20.8}$} & 67.2\textcolor{black}{$_{\uparrow 16.0}$} & \textbf{72.0}\textcolor{black}{$_{\uparrow 15.2}$} & \textbf{68.3}\textcolor{black}{$_{\uparrow 20.0}$} & 82.5\textcolor{black}{$_{\uparrow 29.2}$} & 85.8\textcolor{black}{$_{\uparrow 21.6}$} & \textbf{85.0}\textcolor{black}{$_{\uparrow 21.7}$} & \textbf{79.2}\textcolor{black}{$_{\uparrow 42.5}$} & \textbf{76.5}\textcolor{black}{$_{\uparrow 22.3}$} \\
 & VLM-text & \underline{58.8} & \underline{55.6} & \underline{54.0} & \underline{54.4} & \underline{51.2} & \underline{56.8} & \underline{48.3} & 53.3 & 64.2 & 63.3 & 36.7 & \underline{54.2} \\ \midrule[0.7pt]
\multicolumn{14}{c}{\textit{Open-source general VLMs and pathology VLMs}} \\ \midrule[0.7pt]
\multirow{2}{*}{LLaVA-1.5-7B} & VLM & 40.4\textcolor{black}{$_{\uparrow 5.2}$} & 40.8\textcolor{black}{$_{\uparrow 4.8}$} & 48.8\textcolor{black}{$_{\uparrow 8.8}$} & 41.2\textcolor{black}{$_{\uparrow 9.6}$} & 35.2\textcolor{black}{$_{\uparrow 8.8}$} & 34.0\textcolor{black}{$_{\uparrow 12.4}$} & 38.3\textcolor{black}{$_{\uparrow 4.1}$} & 60.0\textcolor{black}{$_{\uparrow 8.3}$} & 67.5\textcolor{black}{$_{\uparrow 28.3}$} & 71.7\textcolor{black}{$_{\uparrow 13.4}$} & 29.2\textcolor{black}{$_{\uparrow 6.7}$} & 46.1\textcolor{black}{$_{\uparrow 10.0}$} \\
 & VLM-text & 35.2 & 36.0 & 40.0 & 31.6 & 26.4 & 21.6 & 34.2 & 51.7 & 39.2 & 58.3 & 22.5 & 36.1 \\
\multirow{2}{*}{Quilt-LLaVA} & VLM & 46.4\textcolor{black}{$_{\uparrow 8.8}$} & 49.6\textcolor{black}{$_{\uparrow 6.4}$} & 42.4\textcolor{black}{$_{\uparrow 4.0}$} & 49.2\textcolor{black}{$_{\uparrow 9.2}$} & 34.0\textcolor{black}{$_{\uparrow 7.2}$} & 34.8\textcolor{black}{$_{\uparrow 10.0}$} & 32.5\textcolor{black}{$_{\uparrow 1.7}$} & 57.5\textcolor{black}{$_{\uparrow 9.2}$} & 70.8\textcolor{black}{$_{\uparrow 21.6}$} & 71.7\textcolor{black}{$_{\uparrow 14.2}$} & 34.2\textcolor{black}{$_{\uparrow 3.4}$} & 47.6\textcolor{black}{$_{\uparrow 8.7}$} \\
 & VLM-text & 37.6 & 43.2 & 38.4 & 40.0 & 26.8 & 24.8 & 30.8 & 48.3 & 49.2 & 57.5 & 30.8 & 38.9 \\
\multirow{2}{*}{LLaVA-1.5-13B} & VLM & 42.0\textcolor{black}{$_{\uparrow 16.4}$} & 45.6\textcolor{black}{$_{\uparrow 11.2}$} & 44.0\textcolor{black}{$_{\uparrow 10.0}$} & 42.0\textcolor{black}{$_{\uparrow 14.4}$} & 37.6\textcolor{black}{$_{\uparrow 11.2}$} & 37.6\textcolor{black}{$_{\uparrow 11.2}$} & 48.3\textcolor{black}{$_{\uparrow 10.8}$} & 63.3\textcolor{black}{$_{\uparrow 10.0}$} & 71.7\textcolor{black}{$_{\uparrow 18.4}$} & 74.2\textcolor{black}{$_{\uparrow 15.0}$} & 35.8\textcolor{black}{$_{\uparrow 11.6}$} & 49.3\textcolor{black}{$_{\uparrow 12.8}$} \\
 & VLM-text & 25.6 & 34.4 & 34.0 & 27.6 & 26.4 & 26.4 & 37.5 & 53.3 & 53.3 & 59.2 & 24.2 & 36.5 \\
\multirow{2}{*}{PathGen-LLaVA} & VLM & 55.6\textcolor{black}{$_{\uparrow 18.8}$} & 53.2\textcolor{black}{$_{\uparrow 15.6}$} & 56.0\textcolor{black}{$_{\uparrow 13.6}$} & 58.8\textcolor{black}{$_{\uparrow 16.4}$} & 54.8\textcolor{black}{$_{\uparrow 15.2}$} & 60.8\textcolor{black}{$_{\uparrow 17.2}$} & 55.8\textcolor{black}{$_{\uparrow 19.1}$} & 80.0\textcolor{black}{$_{\uparrow 13.3}$} & 80.0\textcolor{black}{$_{\uparrow 17.5}$} & 78.3\textcolor{black}{$_{\uparrow 16.6}$} & 57.5\textcolor{black}{$_{\uparrow 22.5}$} & 62.8\textcolor{black}{$_{\uparrow 16.9}$} \\
 & VLM-text & 36.8 & 37.6 & 42.4 & 42.4 & 39.6 & 43.6 & 36.7 & \underline{66.7} & 62.5 & 61.7 & 35.0 & 45.9 \\
\multirow{2}{*}{Qwen2-VL-7B} & VLM & 40.0\textcolor{black}{$_{\uparrow 14.8}$} & 38.4\textcolor{black}{$_{\uparrow 13.6}$} & 40.4\textcolor{black}{$_{\uparrow 15.6}$} & 41.2\textcolor{black}{$_{\uparrow 17.2}$} & 30.8\textcolor{black}{$_{\uparrow 10.0}$} & 40.4\textcolor{black}{$_{\uparrow 18.8}$} & 51.7\textcolor{black}{$_{\uparrow 16.7}$} & 75.8\textcolor{black}{$_{\uparrow 10.0}$} & 63.3\textcolor{black}{$_{\uparrow 12.5}$} & 77.5\textcolor{black}{$_{\uparrow 22.5}$} & 47.5\textcolor{black}{$_{\uparrow 15.0}$} & 49.7\textcolor{black}{$_{\uparrow 15.1}$} \\
 & VLM-text & 25.2 & 24.8 & 24.8 & 24.0 & 20.8 & 21.6 & 35.0 & 65.8 & 50.8 & 55.0 & 32.5 & 34.6 \\
\multirow{2}{*}{HuatuoGPT-Vision-7B} & VLM & 46.4\textcolor{black}{$_{\uparrow 16.0}$} & 49.6\textcolor{black}{$_{\uparrow 23.2}$} & 48.8\textcolor{black}{$_{\uparrow 20.0}$} & 46.8\textcolor{black}{$_{\uparrow 17.6}$} & 31.6\textcolor{black}{$_{\uparrow 10.4}$} & 40.8\textcolor{black}{$_{\uparrow 10.4}$} & 49.2\textcolor{black}{$_{\uparrow 17.5}$} & 69.2\textcolor{black}{$_{\uparrow 17.5}$} & 75.0\textcolor{black}{$_{\uparrow 26.7}$} & 70.0\textcolor{black}{$_{\uparrow 15.8}$} & 52.5\textcolor{black}{$_{\uparrow 23.3}$} & 52.7\textcolor{black}{$_{\uparrow 18.0}$} \\
 & VLM-text & 30.4 & 26.4 & 28.8 & 29.2 & 21.2 & 30.4 & 31.7 & 51.7 & 48.3 & 54.2 & 29.2 & 34.7 \\
\multirow{2}{*}{Qwen2.5-VL-7B} & VLM & 44.0\textcolor{black}{$_{\uparrow 22.8}$} & 43.2\textcolor{black}{$_{\uparrow 14.8}$} & 42.4\textcolor{black}{$_{\uparrow 16.4}$} & 44.4\textcolor{black}{$_{\uparrow 25.2}$} & 42.4\textcolor{black}{$_{\uparrow 20.8}$} & 44.0\textcolor{black}{$_{\uparrow 24.8}$} & 56.7\textcolor{black}{$_{\uparrow 20.9}$} & 75.0\textcolor{black}{$_{\uparrow 12.5}$} & 75.8\textcolor{black}{$_{\uparrow 10.8}$} & 81.7\textcolor{black}{$_{\uparrow 18.4}$} & 41.7\textcolor{black}{$_{\uparrow 16.7}$} & 53.8\textcolor{black}{$_{\uparrow 18.6}$} \\
 & VLM-text & 21.2 & 28.4 & 26.0 & 19.2 & 21.6 & 19.2 & 35.8 & 62.5 & 65.0 & 63.3 & 25.0 & 35.2 \\
\multirow{2}{*}{Patho-R1-7B} & VLM & 60.4\textcolor{black}{$_{\uparrow 21.6}$} & 52.8\textcolor{black}{$_{\uparrow 17.6}$} & 52.8\textcolor{black}{$_{\uparrow 18.8}$} & 55.2\textcolor{black}{$_{\uparrow 19.6}$} & 60.0\textcolor{black}{$_{\uparrow 21.6}$} & 59.6\textcolor{black}{$_{\uparrow 23.6}$} & 62.5\textcolor{black}{$_{\uparrow 22.5}$} & 78.3\textcolor{black}{$_{\uparrow 15.0}$} & 81.7\textcolor{black}{$_{\uparrow 11.7}$} & 75.0\textcolor{black}{$_{\uparrow 6.7}$} & 60.8\textcolor{black}{$_{\uparrow 19.1}$} & 63.6\textcolor{black}{$_{\uparrow 18.0}$} \\
 & VLM-text & 38.8 & 35.2 & 34.0 & 35.6 & 38.4 & 36.0 & 40.0 & 63.3 & \underline{70.0} & \underline{68.3} & 41.7 & 45.6 \\
\multirow{2}{*}{Lingshu-7B} & VLM & 61.2\textcolor{black}{$_{\uparrow 21.6}$} & 53.2\textcolor{black}{$_{\uparrow 17.2}$} & 50.8\textcolor{black}{$_{\uparrow 18.8}$} & 55.6\textcolor{black}{$_{\uparrow 19.6}$} & 59.2\textcolor{black}{$_{\uparrow 20.0}$} & 59.6\textcolor{black}{$_{\uparrow 24.4}$} & 57.5\textcolor{black}{$_{\uparrow 13.3}$} & 74.2\textcolor{black}{$_{\uparrow 21.7}$} & 86.7\textcolor{black}{$_{\uparrow 21.7}$} & 82.5\textcolor{black}{$_{\uparrow 16.7}$} & 64.2\textcolor{black}{$_{\uparrow 17.5}$} & 64.1\textcolor{black}{$_{\uparrow 19.4}$} \\
 & VLM-text & 39.6 & 36.0 & 32.0 & 36.0 & 39.2 & 35.2 & 44.2 & 52.5 & 65.0 & 65.8 & \underline{46.7} & 44.7 \\ \midrule[0.7pt]
\multicolumn{14}{c}{\textit{More open-source general VLMs}} \\ \midrule[0.7pt]
\multirow{2}{*}{Qwen3-VL-8B} & VLM & 45.2\textcolor{black}{$_{\uparrow 23.6}$} & 44.0\textcolor{black}{$_{\uparrow 16.8}$} & 42.4\textcolor{black}{$_{\uparrow 21.6}$} & 45.6\textcolor{black}{$_{\uparrow 22.8}$} & 36.8\textcolor{black}{$_{\uparrow 14.8}$} & 40.8\textcolor{black}{$_{\uparrow 20.4}$} & 55.8\textcolor{black}{$_{\uparrow 22.5}$} & 74.2\textcolor{black}{$_{\uparrow 25.0}$} & 80.8\textcolor{black}{$_{\uparrow 26.6}$} & 84.2\textcolor{black}{$_{\uparrow 24.2}$} & 55.8\textcolor{black}{$_{\uparrow 25.0}$} & 55.1\textcolor{black}{$_{\uparrow 22.2}$} \\
 & VLM-text & 21.6 & 27.2 & 20.8 & 22.8 & 22.0 & 20.4 & 33.3 & 49.2 & 54.2 & 60.0 & 30.8 & 32.9 \\
\multirow{2}{*}{InternVL3-8B} & VLM & 47.2\textcolor{black}{$_{\uparrow 23.2}$} & 45.6\textcolor{black}{$_{\uparrow 15.2}$} & 44.4\textcolor{black}{$_{\uparrow 18.8}$} & 44.8\textcolor{black}{$_{\uparrow 23.2}$} & 36.4\textcolor{black}{$_{\uparrow 9.6}$} & 41.6\textcolor{black}{$_{\uparrow 19.6}$} & 50.8\textcolor{black}{$_{\uparrow 17.5}$} & 71.7\textcolor{black}{$_{\uparrow 21.7}$} & 72.5\textcolor{black}{$_{\uparrow 20.8}$} & 81.7\textcolor{black}{$_{\uparrow 15.9}$} & 47.5\textcolor{black}{$_{\uparrow 12.5}$} & 53.1\textcolor{black}{$_{\uparrow 18.0}$} \\
 & VLM-text & 24.0 & 30.4 & 25.6 & 21.6 & 26.8 & 22.0 & 33.3 & 50.0 & 51.7 & 65.8 & 35.0 & 35.1 \\
\multirow{2}{*}{Llama-3.2-11B} & VLM & 47.6\textcolor{black}{$_{\uparrow 26.4}$} & 44.0\textcolor{black}{$_{\uparrow 16.0}$} & 38.0\textcolor{black}{$_{\uparrow 14.4}$} & 45.2\textcolor{black}{$_{\uparrow 23.2}$} & 33.2\textcolor{black}{$_{\uparrow 8.4}$} & 40.0\textcolor{black}{$_{\uparrow 20.8}$} & 52.5\textcolor{black}{$_{\uparrow 13.3}$} & 74.2\textcolor{black}{$_{\uparrow 16.7}$} & 76.7\textcolor{black}{$_{\uparrow 21.7}$} & 80.8\textcolor{black}{$_{\uparrow 20.8}$} & 55.0\textcolor{black}{$_{\uparrow 24.2}$} & 53.4\textcolor{black}{$_{\uparrow 18.7}$} \\
 & VLM-text & 21.2 & 28.0 & 23.6 & 22.0 & 24.8 & 19.2 & 39.2 & 57.5 & 55.0 & 60.0 & 30.8 & 34.7 \\
\multirow{2}{*}{Gemma-3-12B} & VLM & 43.6\textcolor{black}{$_{\uparrow 18.4}$} & 42.8\textcolor{black}{$_{\uparrow 20.0}$} & 46.0\textcolor{black}{$_{\uparrow 16.8}$} & 44.8\textcolor{black}{$_{\uparrow 22.4}$} & 42.4\textcolor{black}{$_{\uparrow 15.2}$} & 43.6\textcolor{black}{$_{\uparrow 23.2}$} & 56.7\textcolor{black}{$_{\uparrow 18.4}$} & 76.7\textcolor{black}{$_{\uparrow 15.0}$} & 73.3\textcolor{black}{$_{\uparrow 18.3}$} & 82.5\textcolor{black}{$_{\uparrow 23.3}$} & 53.3\textcolor{black}{$_{\uparrow 24.1}$} & 55.1\textcolor{black}{$_{\uparrow 19.6}$} \\
 & VLM-text & 25.2 & 22.8 & 29.2 & 22.4 & 27.2 & 20.4 & 38.3 & 61.7 & 55.0 & 59.2 & 29.2 & 35.5 \\
\multirow{2}{*}{Gemma-3-27B} & VLM & 47.6\textcolor{black}{$_{\uparrow 20.8}$} & 46.8\textcolor{black}{$_{\uparrow 24.8}$} & 47.2\textcolor{black}{$_{\uparrow 20.8}$} & 50.4\textcolor{black}{$_{\uparrow 24.8}$} & 36.8\textcolor{black}{$_{\uparrow 13.6}$} & 44.4\textcolor{black}{$_{\uparrow 25.2}$} & 57.5\textcolor{black}{$_{\uparrow 19.2}$} & 72.5\textcolor{black}{$_{\uparrow 23.3}$} & 81.7\textcolor{black}{$_{\uparrow 30.0}$} & 82.5\textcolor{black}{$_{\uparrow 20.8}$} & 57.5\textcolor{black}{$_{\uparrow 25.0}$} & 56.8\textcolor{black}{$_{\uparrow 22.6}$} \\
 & VLM-text & 26.8 & 22.0 & 26.4 & 25.6 & 23.2 & 19.2 & 38.3 & 49.2 & 51.7 & 61.7 & 32.5 & 34.2 \\
\multirow{2}{*}{Qwen3-VL-32B} & VLM & 54.0\textcolor{black}{$_{\uparrow 33.2}$} & 50.4\textcolor{black}{$_{\uparrow 26.8}$} & 48.4\textcolor{black}{$_{\uparrow 25.2}$} & 52.0\textcolor{black}{$_{\uparrow 30.0}$} & 39.6\textcolor{black}{$_{\uparrow 19.6}$} & 56.0\textcolor{black}{$_{\uparrow 33.6}$} & 61.7\textcolor{black}{$_{\uparrow 23.4}$} & 82.5\textcolor{black}{$_{\uparrow 22.5}$} & 82.5\textcolor{black}{$_{\uparrow 27.5}$} & 83.3\textcolor{black}{$_{\uparrow 22.5}$} & 56.7\textcolor{black}{$_{\uparrow 20.9}$} & 60.6\textcolor{black}{$_{\uparrow 25.9}$} \\
 & VLM-text & 20.8 & 23.6 & 23.2 & 22.0 & 20.0 & 22.4 & 38.3 & 60.0 & 55.0 & 60.8 & 35.8 & 34.7 \\
\multirow{2}{*}{InternVL3-38B} & VLM & 50.4\textcolor{black}{$_{\uparrow 30.0}$} & 48.0\textcolor{black}{$_{\uparrow 18.0}$} & 40.4\textcolor{black}{$_{\uparrow 15.6}$} & 49.2\textcolor{black}{$_{\uparrow 27.2}$} & 30.8\textcolor{black}{$_{\uparrow 8.8}$} & 45.2\textcolor{black}{$_{\uparrow 24.0}$} & 50.0\textcolor{black}{$_{\uparrow 13.3}$} & \textbf{84.2}\textcolor{black}{$_{\uparrow 17.5}$} & 71.7\textcolor{black}{$_{\uparrow 10.9}$} & 84.2\textcolor{black}{$_{\uparrow 17.5}$} & 50.8\textcolor{black}{$_{\uparrow 21.6}$} & 55.0\textcolor{black}{$_{\uparrow 18.6}$} \\
 & VLM-text & 20.4 & 30.0 & 24.8 & 22.0 & 22.0 & 21.2 & 36.7 & \underline{66.7} & 60.8 & 66.7 & 29.2 & 36.4 \\
\bottomrule[1pt]
\end{tabular}}
\caption{Full evaluations of 18 VLMs on PathBind-VQA and PathBind-PTA across diagnostic dimensions. ``VLM'' denotes evaluating with images, ``VLM-text'' evaluates without images. PathBind-VQA covers six dimensions (D1--D6); PathBind-PTA covers the first five (D1--D5). The highest results of ``VLM'' and ``VLM-text'' per column are highlighted in \textbf{bold} and \underline{underlined}. The numbers with arrows indicate the improvement of ``VLM'' relative to ``VLM-text''.}
\label{table A_PathBind}
\end{table*}

\begin{table}[t!]
\renewcommand{\arraystretch}{1.2}
\renewcommand{\tabcolsep}{8pt}
\centering
\resizebox{\columnwidth}{!}
{
\begin{tabular}{l|ccc|c}
\toprule[1pt]
Model               & Blank & Same subset & Cross subset & Avg. \\ \midrule[0.7pt]
LLaVA-1.5-7B        & 32.6  & 33.1        & 32.6         & 32.8 \\
Quilt-LLaVA         & 35.6  & 34.0        & 34.6         & 34.7 \\
LLaVA-1.5-13B       & 30.9  & 33.1        & 33.3         & 32.4 \\
PathGen-LLaVA       & 51.4  & 57.4        & 56.3         & 55.0 \\
Qwen2-VL-7B          & 9.0   & 34.2        & 36.6         & 26.6 \\
HuatuoGPT-Vision-7B & 41.5  & 42.7        & 44.1         & 42.8 \\
Qwen2.5-VL-7B       & 30.4  & 34.5        & 34.2         & 33.0 \\
Patho-R1-7B         & 47.8  & 44.6        & 43.6         & 45.3 \\ \bottomrule[1pt]
\end{tabular}
}
\caption{Image ablation experiments of four pathology VLMs and their corresponding base models on PathMMU-test-tiny. Three experimental settings are considered: blank images, an image from the same subset, and an image from a different subset. High accuracy under these settings indicates weak sensitivity to the original visual evidence.}
\label{table A2}
\end{table}

\subsection{Image Perturbation Experiments}
\label{appA2}

To further examine whether pathology VLMs rely on the visual content of the input image, we conduct image perturbation experiments on PathMMU-test-tiny. We consider two types of interventions. First, we progressively blur the input image to weaken fine-grained histopathological details. Second, we replace the original image with a blank image, an image sampled from the same subset, or an image sampled from a different subset, while keeping the question and answer options unchanged.

\begin{figure}[t!]
  \includegraphics[width=1.\columnwidth]{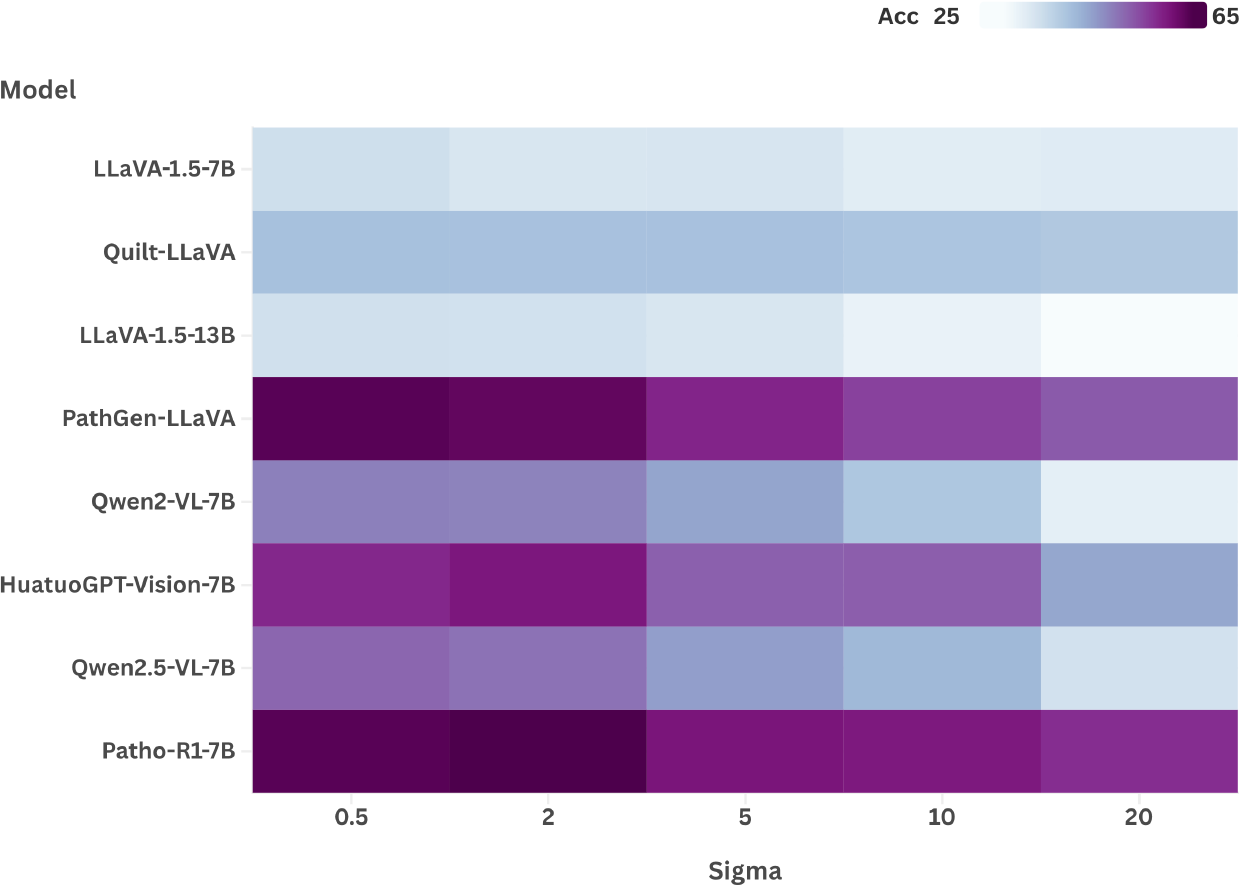}
  \caption{Accuracy heatmap under progressive image blurring on PathMMU-test-tiny. Stable accuracy under strong blur indicates limited sensitivity to fine-grained histopathological details.}
  \label{blur_curve}
\end{figure}

Figure \ref{blur_curve} shows model accuracy under different degrees of image blurring. If a model strongly depends on microscopic visual evidence, its accuracy should decrease as image details become increasingly degraded. However, several models remain relatively stable under moderate or even strong blurring, suggesting limited sensitivity to fine-grained visual evidence. This trend complements the text-only results in Table \ref{table A1}, indicating that benchmark performance can remain high even when the visual signal is weakened.

Table \ref{table A2} further reports results under blank-image and image-replacement settings. Several pathology-tuned models retain substantial accuracy even when the original image is removed or replaced. For example, Patho-R1 achieves 47.8\% accuracy with blank images and remains above 43.0\% under both same-subset and cross-subset image replacement. PathGen-LLaVA shows an even stronger effect, maintaining an average accuracy of 55.0\% across the three ablation settings. These results indicate that some pathology VLMs can preserve many predictions even when the visual evidence is absent or semantically mismatched.

Together, the blur and replacement experiments provide complementary evidence of weak visual dependence. Blurring tests whether models are sensitive to the degradation of image details, while blank-image and image-replacement settings test whether models depend on the correct visual evidence at all. The results suggest that pathology-specific tuning does not necessarily make models more sensitive to visual perturbations, reinforcing the concern that benchmark accuracy may overestimate true visual understanding.

\begin{table}[t!]

\renewcommand{\arraystretch}{1.2}
\renewcommand{\tabcolsep}{8pt}
\centering
\resizebox{\columnwidth}{!}
{
\begin{tabular}{l|cccccc}
\toprule[1pt]
\multirow{2}{*}{Model} & \multicolumn{3}{c}{Test A}                    & \multicolumn{3}{c}{Test B}                  \\ \cline{2-7} 
                       & IoU           & Pre     & Rec        & IoU          & Pre    & Rec        \\ \midrule[0.7pt]
InternVL3-8B           & 7.2           & 10.1          & 25.5          & 3.0          & 3.2          & 29.3          \\
Qwen3-VL-8B            & 15.8          & 20.6          & 55.4          & \textbf{5.9} & \textbf{6.3} & \textbf{59.6} \\
LLaVA-1.5-7B           & 13.2          & 18.2          & 41.8          & 2.4          & 2.6          & 23.2          \\
Quilt-LLaVA            & 10.1          & 15.1          & 30.4          & 2.8          & 3.1          & 26.1          \\
LLaVA-1.5-13B          & 12.4          & 17.2          & 41.5          & 3.6          & 3.8          & 34.1          \\
PathGen-LLaVA          & 15.2          & 20.1          & 50.5          & 4.5          & 4.8          & 42.5          \\
Qwen2-VL-7B            & 11.3          & 15.4          & 40.2          & 3.8          & 4.1          & 37.4          \\
HuatuoGPT-Vision-7B    & 14.0          & 18.7          & 43.9          & 4.4          & 4.7          & 41.4          \\
Qwen2.5-VL-7B           & \textbf{18.6} & \textbf{23.9} & \textbf{61.5} & 5.3          & 5.7          & 55.3          \\
Patho-R1-7B            & 14.9          & 19.9          & 51.0          & 5.1          & 5.5          & 52.1          \\ \bottomrule[1pt]
\end{tabular}
}
\caption{Full attention-grounding results on PathVG. We report the IoU, Precision, and Recall between entity-token attention maps and ground-truth bounding boxes for pathological expressions. The best results are highlighted in \textbf{bold}.}
\label{table A3}
\end{table}

\subsection{Full Grounding Results on PathVG}
\label{appA3}
To complement the grounding analysis in the main paper, we report the complete PathVG results in Table \ref{table A3}. For each model, we compute the overlap between entity-token attention maps and ground-truth bounding boxes using IoU, Precision, and Recall. Results are reported separately on Test A and Test B. Detailed attention extraction and metric definitions are provided in Appendix \ref{appC}.

Overall, grounding performance remains limited across models. Although several models achieve reasonable recall, their IoU and Precision are consistently low, indicating that attention often spreads beyond the annotated pathological regions. For example, Qwen2.5-VL obtains the best IoU on Test A, but its IoU remains only 18.6\%, despite a recall of 61.5\%. On Test B, all models show substantially lower IoU and Precision, suggesting that entity-level localization remains challenging under more difficult grounding scenarios.

The results also show that pathology-specific tuning does not consistently improve grounding. Patho-R1 achieves strong VQA accuracy in the main experiments, but its PathVG IoU is lower than Qwen2.5-VL on both Test A and Test B. Similarly, Quilt-LLaVA underperforms LLaVA-1.5-7B on Test A, although it shows small gains on Test B. These results support the main paper’s conclusion that domain training can improve answer prediction without proportionally strengthening entity-level visual-semantic binding.

\section{Benchmark Construction Details}
\label{appB}
All three PathBind components, including PathBind-PTA, are publicly released via Hugging Face gated access.
\subsection{Details of VQA Sample Filtering}
\label{appB1}
To construct the visual-dependence-enriched VQA component of PathBind, we filter existing pathology VQA samples using a two-step text-only probing pipeline. The goal is to remove samples that can be reliably answered without visual evidence, thereby reducing textual shortcuts and question-option priors before evaluating visual dependence.

In Step 1, we evaluate two closed-source LLMs under the text-only setting and remove samples that both models answer correctly. As shown in Table \ref{table A4}, both models obtain non-trivial text-only accuracy across the source benchmarks, with Gemini-3-Pro reaching an average accuracy of 53.9\% and GPT-5.4 reaching 47.1\%. This confirms that many samples in existing pathology VQA benchmarks can be solved by strong language models without image input.

In Step 2, we apply five diverse LLMs to the remaining samples after Step 1 filtering. We then remove samples that are answered correctly by three or more of these models. This consensus-based filtering further reduces samples that remain text-solvable across diverse model families. After the two filtering steps, the VQA pool is reduced from the original collection to 6,356 samples.

Table \ref{table A5} provides a detailed breakdown of the number of models that answer each sample correctly during the two filtering steps. In Step 1, 35.2\% of samples are answered correctly by both closed-source LLMs on average and are therefore removed. After this filtering, 8,590 samples remain. In Step 2, an additional portion of samples is removed when at least three of the five LLMs answer them correctly.

This two-step filtering strategy is intentionally conservative. The first step removes samples that are easy for strong closed-source LLMs, while the second step removes samples that are consistently solvable by multiple additional LLMs. By combining strong-model filtering and model-consensus filtering, PathBind reduces the influence of text-only shortcuts while preserving a sufficiently large and diverse VQA set for evaluation.

\begin{table*}[t!]
\renewcommand{\arraystretch}{1.2}
\renewcommand{\tabcolsep}{8pt}
\centering
\resizebox{\textwidth}{!}
{
\begin{tabular}{l|clcccc|c}
\toprule[1pt]
Model            & \multicolumn{2}{c}{PathMMU-Test-all} & Path-VQA & Quilt-VQA & MedXpert-Path & OmniMed-Bright & \multicolumn{1}{l}{Avg.} \\ \midrule[0.7pt]
\multicolumn{8}{c}{\textit{Step 1: Filter by two closed-source LLMs}}                                                                                                                 \\ \midrule[0.7pt]
GPT-5.4          & \multicolumn{2}{c}{47.0}             & 47.7     & 42.5      & \textbf{54.5}          & 44.0           & \multicolumn{1}{l}{47.1} \\
Gemini-3-Pro     & \multicolumn{2}{c}{\textbf{50.5}}             & \textbf{59.7}     & \textbf{48.7}      & 54.4          & \textbf{56.1}           & \multicolumn{1}{l}{\textbf{53.9}} \\ \midrule[0.7pt]
\multicolumn{8}{c}{\textit{Step 2: Filter by five additional LLMs}}                                                                                                                 \\ \midrule[0.7pt]
Qwen3.6-Max      & \multicolumn{2}{c}{29.5}             & 21.6     & 12.7      & 20.8          & 39.8           & 24.9                     \\
GLM-5            & \multicolumn{2}{c}{30.1}             & 21.1     & 12.7      & 17.0          & 41.5           & 24.5                     \\
Kimi-K2.6        & \multicolumn{2}{c}{\textbf{35.6}}             & 23.4     & 14.0      & \textbf{22.6}          & \textbf{49.4}           & \textbf{29.0}                     \\
DeepSeek-V4-Pro   & \multicolumn{2}{c}{30.0}             & 22.5     & 13.6      & 20.8          & 31.3           & 23.6                     \\
Llama-4-Maverick & \multicolumn{2}{c}{25.4}             & \textbf{42.7}     & \textbf{14.5}      & 17.0          & 34.9           & 26.9                     \\ 
\bottomrule[1pt]
\end{tabular}
}
\caption{A coarse filtering pipeline using two closed-source LLMs and five additional LLMs on five popular pathology VQA benchmarks. Step 1 reports the accuracy of two closed-source LLMs under the “text-only” setting, then removes samples that both models answer correctly. Step 2 reports the accuracy of five additional LLMs under the “text-only” setting on the remaining samples after Step 1 filtering, then removes samples that are answered correctly by three or more models. The best results in each step are highlighted in \textbf{bold}.}
\label{table A4}
\end{table*}

\begin{table*}[t!]
\renewcommand{\arraystretch}{1.2}
\renewcommand{\tabcolsep}{8pt}
\centering
\resizebox{\textwidth}{!}
{
\begin{tabular}{l|ccccc|c}
\toprule[1pt]
Model        & PathMMU-Test-all & Path-VQA & Quilt-VQA & MedXpert-Path & OmniMed-Bright & Avg. \\ \midrule[0.7pt]
\multicolumn{7}{c}{\textit{Step 1: Filter by two closed-source LLMs}}                                                                                                             \\ \midrule[0.7pt]
0/2 correct & 36.9             & 29.0     & 44.3      & 32.2          & 28.4           & 34.2 \\ 

1/2 correct & 28.7             & 34.5     & 20.1      & 26.7          & 43.0           & 30.6 \\ 

2/2 correct & 34.4             & 36.5     & 35.6      & 41.1          & 28.5           & 35.2 \\ \midrule[0.7pt]
\multicolumn{7}{c}{\textit{Step 2: Filter by five additional LLMs}}                                                                                                             \\ \midrule[0.7pt]
0/5 correct  & 41.8             & 40.1     & 81.9      & 56.6          & 19.3           & 47.9                     \\
1/5 correct  & 18.3             & 33.0     & 4.1       & 11.3          & 24.1           & 18.2                     \\
2/5 correct  & 12.6             & 5.4      & 1.4       & 17.0          & 21.2           & 11.5                     \\
3/5 correct  & 10.2             & 4.5      & 0.0       & 9.4           & 16.7           & 8.2                      \\
4/5 correct  & 8.8              & 11.4     & 2.7       & 3.8           & 13.2           & 8.0                      \\
5/5 correct  & 8.2              & 5.7      & 10.0      & 1.9           & 5.5            & 6.3                      \\ \bottomrule[1pt]
\end{tabular}
}
\caption{Distribution of samples answered correctly by different numbers of text-only models during VQA filtering. After Step 1, samples answered correctly by both closed-source LLMs are removed, leaving 8,590 samples. After Step 2, samples answered correctly by at least three of five LLMs are removed, leaving 6,356 samples.}
\label{table A5}
\end{table*}

\subsection{Details of Grounding Sample Filtering}
\label{appB2}
To construct the grounding component of PathBind, we start from the full PathVG testA and testB splits, which contain 3,048 expressions in total, including 1,342 expressions from testA and 1,706 expressions from testB. Since some expressions are too generic to uniquely identify one bounded region in a histology image, we apply a specificity filtering step before including them in PathBind.

We use GPT-5.4 and Gemini to judge whether each expression is sufficiently specific for region-level grounding. An expression is retained if it identifies a localized pathological entity or visual pattern that can be associated with one bounded region; otherwise, it is removed as vague. 

Two models classify 2,467 expressions as \textsc{Specific}, including 1,047 from testA and 1,420 from testB, and classify 581 expressions as \textsc{Vague}. The resulting retention rate is 80.9\%. This rate is higher than the retention rate of the VQA filtering stage, which is expected because grounding expressions in PathVG are generally more visually specific than VQA questions.

\subsection{PathBind-PTA VQA Construction}
\label{appB3}
The private component of PathBind is constructed from a pathology teaching atlas to provide additional expert-oriented evaluation samples. Unlike the public VQA component, which focuses on removing text-solvable shortcuts from existing benchmarks, the private set is designed to complement public data with carefully controlled pathology concepts and image-question pairs.

We first select high-quality pathology images covering diverse diagnostic scenarios from the teaching atlas. GPT-5.4 is then used to generate initial multiple-choice questions based on the visual content, with each question associated with one of the predefined pathology reasoning dimensions. This process produces 900 candidate image-question pairs.

The pathology experts subsequently review all generated candidates to remove ambiguous cases, text-leakage samples, and questions whose answers cannot be sufficiently supported by visual evidence. The experts also balance the distribution of diagnostic dimensions and answer choices, and add necessary anatomical or clinical context when the original question is under-specified.

After expert refinement, we obtain PathBind-PTA, consisting of 600 multiple-choice questions evenly distributed across five diagnostic dimensions: coarse tissue/organ recognition, cellular morphology, cell-stroma interaction, spatial localization, and diagnostic reasoning. IHC and staining interpretation are excluded because the teaching atlas does not provide sufficient controlled samples for this category.

\subsection{Diagnostic Dimension Definitions}
\label{appB4}

PathBind organizes VQA samples into six diagnostic dimensions. These dimensions are designed to reflect different levels of pathology visual reasoning, from global tissue recognition to fine-grained diagnostic interpretation. Detailed explanations are listed in Table \ref{tab:dimension_definitions}.

\begin{table*}[t]
\centering
\begin{tabular}{p{0.08\linewidth}p{0.28\linewidth}p{0.55\linewidth}}
\toprule[1pt]
Dim. & Name & Definition \\
\midrule[0.7pt]
D1 & Coarse tissue/organ recognition & Recognizing the broad tissue type, organ context, or general histological category from global visual patterns. \\
D2 & Cellular morphology & Identifying cell-level features such as nuclear atypia, cytoplasmic appearance, mitotic figures, necrosis, or cell shape. \\
D3 & Cell--stroma interaction & Interpreting the relationship between cells and surrounding stroma, including infiltration patterns, inflammatory distribution, desmoplasia, and tumor-microenvironment interactions. \\
D4 & Spatial localization & Reasoning about the spatial position, arrangement, or distribution of pathological structures within the image. \\
D5 & Diagnostic reasoning & Integrating visual evidence to support diagnosis, differential diagnosis, grading, staging, or clinically oriented interpretation. \\
D6 & Staining and IHC interpretation & Interpreting staining patterns, immunohistochemical positivity, intensity, distribution, or quantitative staining-related information. \\
\bottomrule[1pt]
\end{tabular}
\caption{Definitions of the six diagnostic dimensions used in PathBind.}
\label{tab:dimension_definitions}
\end{table*}

\subsection{Manual Review Details}
\label{appB5}

\paragraph{Reviewer qualifications and review protocol.}
The expert panel consisted of two attending pathologists and one senior consultant pathologist. The two attending pathologists had 6 and 9 years of diagnostic pathology experience, respectively, while the senior consultant pathologist had 21 years of experience. For all three PathBind components, the two attending pathologists independently reviewed each candidate using predefined task-specific criteria. The reviewers were blinded to model identities, model predictions, attention maps, and downstream evaluation results. Cases with discrepant initial decisions were jointly discussed with the senior consultant pathologist until a final consensus decision was reached.

\paragraph{Inter-rater agreement.}
To assess the reliability of expert curation, inter-rater agreement was calculated from the initial independent retain/exclude decisions of the two attending pathologists before consensus discussion. As summarized in Table~\ref{tab:reviewer_agreement}, the percentage agreement was 94.18\% for PathBind-VQA, 92.91\% for PathBind-Grounding, and 96.44\% for PathBind-PTA, with corresponding Cohen's $\kappa$ values of 0.851, 0.858, and 0.912, respectively. These results indicate consistently high inter-rater agreement across the three curation tasks. Discrepant cases were subsequently discussed with the senior consultant pathologist until a final consensus decision was reached. The final benchmark sizes were determined after consensus resolution and, where applicable, subsequent quality ranking and distribution balancing.

We provide additional details of the expert review process used in PathBind construction. Expert review is applied after automated filtering for all three components of PathBind, but the review criteria differ according to the type of data: VQA samples are reviewed for visual dependence and question quality, grounding samples are reviewed for expression-bbox consistency, and private teaching atlas samples are reviewed for visual support and answer validity.

\begin{table*}[t]
\centering
\renewcommand{\arraystretch}{1.15}
\renewcommand{\tabcolsep}{6pt}
\resizebox{\textwidth}{!}{
\begin{tabular}{lccccccc}
\toprule[1pt]
Component
& Reviewed
& Both retain
& Both exclude
& R1 retain/R2 exclude
& R1 exclude/R2 retain
& Agreement (\%)
& Cohen's $\kappa$ \\
\midrule[0.7pt]
PathBind-VQA
& 6,356
& 1,510
& 4,476
& 185
& 185
& 94.18
& 0.851 \\

PathBind-Grounding
& 2,467
& 1,146
& 1,146
& 87
& 88
& 92.91
& 0.858 \\

PathBind-PTA
& 900
& 630
& 238
& 15
& 17
& 96.44
& 0.912 \\
\bottomrule[1pt]
\end{tabular}}
\caption{Inter-rater agreement for expert review of the three PathBind components.
Agreement was calculated from the initial independent retain/exclude decisions of the two attending pathologists before consensus discussion.}
\label{tab:reviewer_agreement}
\end{table*}

\paragraph{PathBind-VQA review.}
For the public VQA component, the two attending pathologists independently reviewed the 6,356 candidates retained after two-stage text-only filtering. In the first stage of manual review, samples were excluded if they contained low-information images, image--question mismatch, under-specified questions, redundant templates, text leakage within images, or multi-panel composites. Specifically, 40 low-information samples, 15 image--question mismatch samples, 126 under-specified samples, 2,125 redundant or near-duplicate templates, 130 samples with text leakage, and 129 multi-panel composite samples were removed, leaving 3,791 eligible candidates. In the second stage, the remaining candidates were comparatively assessed within each diagnostic dimension according to image clarity, question specificity, clinical relevance, and the plausibility and discriminative quality of the answer options. Discrepant assessments were resolved through consensus discussion with the senior consultant pathologist. The highest-quality samples were then selected while maintaining balanced coverage across diagnostic dimensions and answer choices, yielding 1,500 VQA samples, with 250 samples in each of the six dimensions.

\paragraph{PathBind-Grounding review.}
For the grounding component, the two attending pathologists independently reviewed the 2,467 candidates retained from the initial 3,048 PathVG samples after expression-specificity filtering. In the first stage of manual review, each expression-bounding-box triplet was assessed for spatial-description accuracy, content-description accuracy, and bounding-box coverage of the referred pathological region. The review excluded 210 samples with inaccurate spatial descriptions, 60 with inaccurate content descriptions, and 970 with inaccurate bounding boxes, leaving 1,227 eligible candidates. In the second stage, the remaining candidates were comparatively evaluated according to image clarity, expression specificity, expression-region correspondence, bounding-box accuracy, and the representativeness of the pathological findings. Disagreements were resolved through consensus discussion with the senior consultant pathologist. Based on this quality-oriented selection, 500 region-level grounding samples were retained, including 380 samples from testA and 120 samples from testB.

\paragraph{PathBind-PTA review.}
For the private teaching atlas component, the two attending pathologists independently review the 900 generated VQA candidates. The review removes ambiguous questions, text-leakage candidates, and questions whose answers are not sufficiently supported by the image. The experts also add necessary anatomical or system context when the diagnosis would otherwise be under-specified. Since staining and IHC questions are more likely to depend on textual labels or external staining protocols, this dimension is excluded from the private set. The final PathBind-PTA component contains 600 multiple-choice questions across the first five diagnostic dimensions, with 120 samples per dimension. We also balance answer choices, resulting in 150 samples for each option A-D. Across all three components, discrepant retain/exclude decisions and disagreements regarding exclusion reasons were discussed with the senior consultant pathologist, and the final annotation was determined by consensus.

\section{Attention-Based Grounding Metrics}
\label{appC}

\subsection{Attention Extraction}
\label{appC1}

We follow the token-wise attention extraction strategy inspired by HiDe \cite{liu2025hide}, which uses the attention weights of informative text tokens to identify their corresponding visual regions. In our setting, the informative text tokens are the tokens of a pathological entity expression. Given an image and an entity phrase, we extract the attention from the entity tokens to image patches, aggregate the attention maps across heads, layers, and entity tokens, and convert the resulting heatmap into a binary attended region.

Let the input sequence contain a contiguous image-token span $[s_{\mathrm{img}}, e_{\mathrm{img}})$ with $N_{\mathrm{img}}=e_{\mathrm{img}}-s_{\mathrm{img}}$ image tokens. The image-token span is identified according to the model family. For Qwen-VL-style models, we locate visual tokens between the vision start and end tokens and reshape them according to the visual grid. For LLaVA-style models, the image placeholder is expanded into a contiguous block of visual tokens. Other models are handled analogously according to their processor-defined image-token layout. Let $\mathcal{K}$ denote the token positions corresponding to the queried pathological entity phrase.

For each entity-token position $k\in\mathcal{K}$ and decoder layer $\ell$, we extract the attention from the token $k$ to all image tokens and average it over attention heads:
\begin{equation}
    a_{\ell, k}=\frac{1}{H} \sum_{h=1}^{H} \operatorname{Attn}_{\ell, h}\left[k, s_{\mathrm{img}}: e_{\mathrm{img}}\right] \in \mathbb{R}^{N_{\mathrm{img}}},
\end{equation}
where $H$ is the number of attention heads. This produces one patch-level attention vector for each entity token and layer.

\paragraph{Layer selection.}
For all evaluated models, we aggregate entity-token attention over the middle third of decoder layers,
\begin{equation}
    \mathcal{L}_{\text{mid}} = \left\{ \left\lfloor \frac{L}{3} \right\rfloor, \left\lfloor \frac{L}{3} \right\rfloor + 1, \dots, \left\lfloor \frac{2L}{3} \right\rfloor - 1 \right\},
\end{equation}
where $L$ is the number of language-decoder layers. This choice is motivated by the empirical observation from HiDe, which scans the average semantic-token attention on ground-truth regions across every decoder layer of Qwen2.5-VL and InternVL3, and shows that mid-decoder layers provide the strongest region signal. HiDe's public code selects a single peak layer per architecture; both of these peak layers lie inside the middle third of their respective architectures, so our middle-third average includes each of HiDe's chosen layers as a member.

To rule out a domain- or architecture-transfer failure of this default, we conduct a sanity check on six representative models spanning the LLaVA-1.5, InternVL3, and Qwen families on PathBind-Grounding. Using the per-layer entity-max maps saved during the main-table run, Table~\ref{tab:sweep_layers} compares mean IoU under four aggregation strategies: the shallow-third, middle-third (default), deep-third, and all-layer means. Middle-third aggregation is the best-performing choice on five of the six models, and no pathology-tuned model shifts its peak grounding layer outside the middle third relative to its general base. InternVL3-8B is the only exception, achieving a marginally higher IoU (+1.9 points) with deep-third aggregation; we retain middle-third as the paper-wide default to keep the aggregation protocol architecture-agnostic.

\begin{table}[t!]
\centering
\resizebox{\columnwidth}{!}{
\begin{tabular}{l|cccc}
\toprule
Model & Shallow-third & Mid-third & Deep-third & All-layers \\
\midrule
LLaVA-1.5-7B     & 5.8 & \textbf{14.1} & 10.9 & 10.2 \\
Quilt-LLaVA      & 5.6 & \textbf{10.4} & 8.6 & 8.1 \\
InternVL3-8B     & 4.4 & 8.5 & \textbf{10.4} & 8.0 \\
Qwen3-VL-8B      & 7.2 & \textbf{18.8} & 11.9 & 15.7 \\
Qwen2.5-VL-7B    & 7.3 & \textbf{19.7} & 16.5 & 17.7 \\
Patho-R1-7B      & 7.8 & \textbf{15.0} & 10.9 & 12.5 \\
\bottomrule
\end{tabular}}
\caption{Layer-aggregation sanity check on PathBind-Grounding. Mean IoU (\%) under four layer-aggregation strategies. Best value per row is in \textbf{bold}.}
\label{tab:sweep_layers}
\end{table}

For each entity token $k$, we then average the head-pooled attention vectors across the selected layers:
\begin{equation}
    m_k = \frac{1}{|\mathcal{L}_{\text{mid}}|} \sum_{\ell \in \mathcal{L}_{\text{mid}}} a_{\ell,k}.
\end{equation}
The resulting vector $m_k$ is then reshaped into a two-dimensional patch grid of size $H_p\times W_p$ according to the model-specific visual-token layout.

When an entity phrase contains multiple tokens, we aggregate their maps using element-wise maximum:
\begin{equation}
    M(i,j) = \max_{k \in \mathcal{K}} m_k(i,j).
\end{equation}
We use maximum pooling rather than average pooling so that a strongly localized token can contribute to the final entity map without being diluted by other tokens in the phrase.

The patch-level map $M$ is then smoothed with a Gaussian kernel with $\sigma=3$, min-max normalized to $[0,1]$, and bilinearly upsampled to the original image resolution. We denote the final image-resolution heatmap as $\widehat{M}$. To obtain a binary attended region, we select the top $20\%$ pixels by attention value:
\begin{equation}
    \theta_q = \text{Quantile}(\widehat{M}, 1-q), \quad \text{Mask}(u) = \mathbb{1}[\widehat{M}(u) \ge \theta_q].
\end{equation}
This thresholding strategy fixes the attended area ratio across models and samples, making region-overlap metrics directly comparable.

\paragraph{Choice of $q=0.2$.}
Unlike HiDe's fixed-value threshold $\alpha \in [0,1]$ applied to the normalized heatmap, which yields per-model attended areas of vastly different sizes depending on how peaky each model's attention distribution is. Using a retained-area fraction \(q\), our quantile formulation fixes the attended-area ratio to \(q\) across all models, ensuring that IoU, Precision, and Recall are directly comparable across architectures. We set $q=0.2$ to approximately match the 90th percentile of the ground-truth bounding-box area distribution in PathBind-Grounding (median $=4.8\%$, 90th percentile $=20.2\%$), giving Recall enough budget without diluting Precision below the level where absolute differences remain interpretable. Empirically, peak IoU for the four highest-scoring models (LLaVA-1.5-7B, Qwen3-VL-8B, Qwen2.5-VL-7B, Patho-R1-7B) lies in $q \in [0.1,0.2]$ (Table~\ref{tab:sweep_thr}); the remaining two (Quilt-LLaVA and InternVL3-8B) increase near-monotonically up to $q=0.4$ but with modest differences. $q=0.2$ is therefore at the peak for the strongest models and within 1.1 points of the peak for all others, providing a fair operating point across attention distributions.

\paragraph{Sensitivity ablation.}
To verify that the model ranking induced by this choice is not an artefact of any specific threshold, we sweep $q \in \{0.05,0.1,0.2,0.3,0.4\}$ and additionally report a threshold-free pixel-level Average Precision (AP), which is the area under the precision-recall curve where each image pixel is scored by its attention value and labelled by the ground-truth bounding box. Layer aggregation follows the default middle-third mean throughout. Table~\ref{tab:sweep_thr} reports these values on six representative models on PathBind-Grounding. Every pairwise Kendall's $\tau$ between the five threshold-based rankings and the threshold-free AP ranking equals 1.0, indicating that the ordering of the six evaluated models remains invariant across the tested thresholds.

\begin{table}[t!]
\centering
\resizebox{\columnwidth}{!}{
\begin{tabular}{l|ccccc|c}
\toprule
Model & IoU$_{q=0.05}$ & IoU$_{q=0.1}$ & IoU$_{q=0.2}$
      & IoU$_{q=0.3}$ & IoU$_{q=0.4}$ & AP \\
\midrule
LLaVA-1.5-7B     & 11.8 & 13.7 & 14.1 & 13.4 & 12.6 & 28.4 \\
Quilt-LLaVA      & 6.8 & 8.8 & 10.4 & 10.9 & 11.0 & 20.7 \\
InternVL3-8B     & 6.2 & 7.7 & 8.5 & 9.1 & 9.6 & 19.2 \\
Qwen3-VL-8B      & 20.8 & 21.3 & 18.8 & 16.5 & 14.6 & 41.5 \\
Qwen2.5-VL-7B    & \textbf{21.1} & \textbf{23.0} & \textbf{19.7} & \textbf{16.6} & \textbf{14.7} & \textbf{42.5} \\
Patho-R1-7B      & 15.8 & 16.1 & 15.0 & 14.1 & 13.2 & 34.1 \\
\bottomrule
\end{tabular}}
\caption{Threshold sensitivity + pixel-AP on PathBind-Grounding. Mean IoU (\%) at each threshold and pixel-level Average Precision (AP, \%) computed with mid-third layer aggregation. All pairwise Kendall's $\tau$ values among the six columns are 1.00, i.e., the model ranking is identical across all thresholds and AP. Best value per column is in \textbf{bold}.}
\label{tab:sweep_thr}
\end{table}

The overall procedure is summarized in Algorithm~\ref{alg:attention_extraction}. 
Unless otherwise specified, we use the default hyperparameters: middle-third layer aggregation, Gaussian smoothing with $\sigma=3$, and top-$20\%$ thresholding.

\begin{algorithm}[h!]
\caption{Entity-token attention extraction}
\label{alg:attention_extraction}
\begin{algorithmic}[1]
\REQUIRE Input image $I$, prompt $q$, entity phrase $e$, model $f$
\ENSURE Binary attended region $\mathrm{Mask}$
\STATE Run a forward pass with attention outputs enabled.
\STATE Identify the image-token span $[s_{\mathrm{img}}, e_{\mathrm{img}})$.
\STATE Locate entity-token positions $\mathcal{K}$ corresponding to phrase $e$.
\STATE Select middle-third decoder layers $\mathcal{L}_{\mathrm{mid}}$.
\FOR{$k \in \mathcal{K}$}
\FOR{$\ell \in \mathcal{L}_{\mathrm{mid}}$}
\STATE Extract attention from token $k$ to image tokens.
\STATE Average the attention over heads to obtain $a_{\ell,k}$.
\ENDFOR
\STATE Average $\{a_{\ell,k}\}_{\ell\in\mathcal{L}_{\mathrm{mid}}}$ over layers to obtain $m_k$.
\STATE Reshape $m_k$ into a patch grid of size $H_p\times W_p$.
\ENDFOR
\STATE Aggregate entity-token maps by element-wise maximum to obtain $M$.
\STATE Smooth $M$ with Gaussian kernel $\sigma=3$.
\STATE Min-max normalize and upsample $M$ to the original image resolution, yielding $\widehat{M}$.
\STATE Threshold $\widehat{M}$ at the top $20\%$ pixels to obtain $\mathrm{Mask}$.
\RETURN $\mathrm{Mask}$
\end{algorithmic}
\end{algorithm}

\subsection{Region-Overlap Metrics}
\label{appC2}
After obtaining the binary attended region $\mathrm{Mask}$ from the entity-token attention heatmap, we compare it with the ground-truth bounding box of the corresponding pathological expression. The ground-truth box is converted into a binary region mask $\mathrm{GT}$ at the original image resolution, where pixels inside the annotated box are assigned $1$, and all other pixels are assigned $0$.

Let $A = |\mathrm{Mask}\cap \mathrm{GT}|, 
U = |\mathrm{Mask}\cup \mathrm{GT}|,
M = |\mathrm{Mask}|,
G = |\mathrm{GT}|.$ Here, $A$ denotes the intersection area between the attended region and the ground-truth region, $U$ denotes their union area, $M$ denotes the attended area, and $G$ denotes the ground-truth area. We then compute three region-overlap metrics:

\begin{equation}
    \mathrm{IoU} = \frac{A}{U},
\quad
\mathrm{Precision} = \frac{A}{M},
\quad
\mathrm{Recall} = \frac{A}{G}.
\end{equation}

IoU measures the overall spatial agreement between the attention-derived region and the ground-truth region. Precision measures how much of the attended region falls inside the ground-truth box, reflecting whether the model attends specifically to the annotated pathological entity. Recall measures how much of the ground-truth box is covered by the attended region, reflecting whether the model touches the target entity at all.

These metrics capture complementary aspects of attention grounding. A model with high Recall but low Precision or IoU may attend to the correct region but also spread attention broadly outside it. In contrast, high Precision indicates more selective attention, while high IoU requires both sufficient coverage of the target region and limited attention leakage outside the target. Therefore, we report all three metrics in Table \ref{table A3} to distinguish loose attention overlap from precise entity-level visual grounding.

\subsection{Effective Attention Support}
\label{appC3}
In addition to region-overlap metrics, we quantify how diffuse the entity-token attention is over image patches. Although normalized entropy can measure attention diffuseness, its values are often close to $1$ across models and are therefore less intuitive to interpret. We instead report the effective attention support ratio, which converts entropy into the equivalent number of uniformly attended patches.

Let $p=(p_1,\ldots,p_N)$ be the normalized attention distribution over $N=H_p\times W_p$ image patches, where $\sum_i p_i=1$. The Shannon entropy of the distribution is
\begin{equation}
    H(p) = -\sum_{i=1}^{N} p_i \log p_i.
\end{equation}
We define the effective attention support size as
\begin{equation}
    N_{\text{eff}} = \exp(H(p)).
\end{equation}
Equivalently, if $\tilde{H}=H(p)/\log N$ is the normalized entropy, then
\begin{equation}
    N_{\text{eff}} = N^{\widetilde{H}}.
\end{equation}

The quantity $N_{\mathrm{eff}}$ can be interpreted as the number of patches in a uniform distribution with the same entropy as the observed attention distribution. For example, if all attention mass is concentrated on one patch, then $H(p)=0$ and $N_{\mathrm{eff}}=1$. If attention is uniformly distributed over all patches, then $H(p)=\log N$ and $N_{\mathrm{eff}}=N$. Intermediate values indicate partial concentration.

We report the normalized ratio $N_{\mathrm{eff}}/N$ averaged over samples. A value close to 1 indicates that the observed attention distribution has entropy close to that of a uniform distribution over all image patches, whereas a smaller value indicates stronger concentration. Importantly, $N_{\mathrm{eff}}$ is an entropy-equivalent support size and does not imply that exactly $N_{\mathrm{eff}}$ patches receive identical attention weights. Figure \ref{fig2} (b) shows near-full effective support across models, suggesting weak spatial concentration of entity-token attention.

It is a well-known property of standard scaled dot-product attention that softmax naturally disperses probability mass over correlated visual features. We therefore do \emph{not} interpret $N_{\mathrm{eff}}/N$ as an absolute measure of grounding failure; a high value alone does not imply that the model lacks visual grounding. Instead we interpret $N_{\mathrm{eff}}/N$ strictly through cross-model contrasts on the same
PathBind-Grounding samples: (i) it is uniformly high across all evaluated pathology VLMs, indicating that no model produces a compact entity-specific attention peak on our benchmark; (ii) pathology tuning does not measurably tighten $N_{\mathrm{eff}}/N$ within any  pair; (iii) our conclusion is therefore about the relative failure of pathology
tuning to sharpen entity-selectivity, not about attention being diffuse in an absolute sense.

\subsection{Cross-Query Attention Correlation}
\label{appC4}
Effective attention support measures whether entity-token attention is spatially concentrated, but it does not test whether the attention map changes according to the queried entity. To evaluate query specificity, we compute cross-query attention correlation on the same image using different pathological entity queries.

For each sample, we construct $K=3$ entity queries: the ground-truth entity expression of the current sample and two alternative entity expressions sampled from other examples in the same split. The alternative expressions are constrained to be different from the current entity expression. For each query, we extract an attention heatmap over the same image using the procedure described above. This results in $K$ heatmaps, denoted as ${\mathrm{heat}_1,\ldots,\mathrm{heat}_K}$.

We then compute the Pearson correlation between every pair of heatmaps and average the correlations:
\begin{equation}
    B_2(\text{sample}) = \frac{1}{(\frac{K}{2})} \sum_{a<b} \text{Pearson}(\text{heat}_a, \text{heat}_b).
\end{equation}
Since $K=3$ in our experiments, each sample contains three pairwise comparisons.

A high cross-query attention correlation indicates that different pathological entity queries produce similar attention patterns on the same image. This suggests that the model’s visual evidence is weakly conditioned on the queried entity. In contrast, a lower correlation indicates that attention changes more substantially when the target entity changes, suggesting stronger query-specific grounding.

To ensure cross-model comparability, the alternative entity queries are sampled with a fixed seed for each sample and are kept identical across models. We report the mean cross-query attention correlation over samples in Figure \ref{fig2} (c).

\subsection{Qualitative Attention Visualization}
\label{appC5}

\begin{figure*}[t!]
  \includegraphics[width=1.\textwidth]{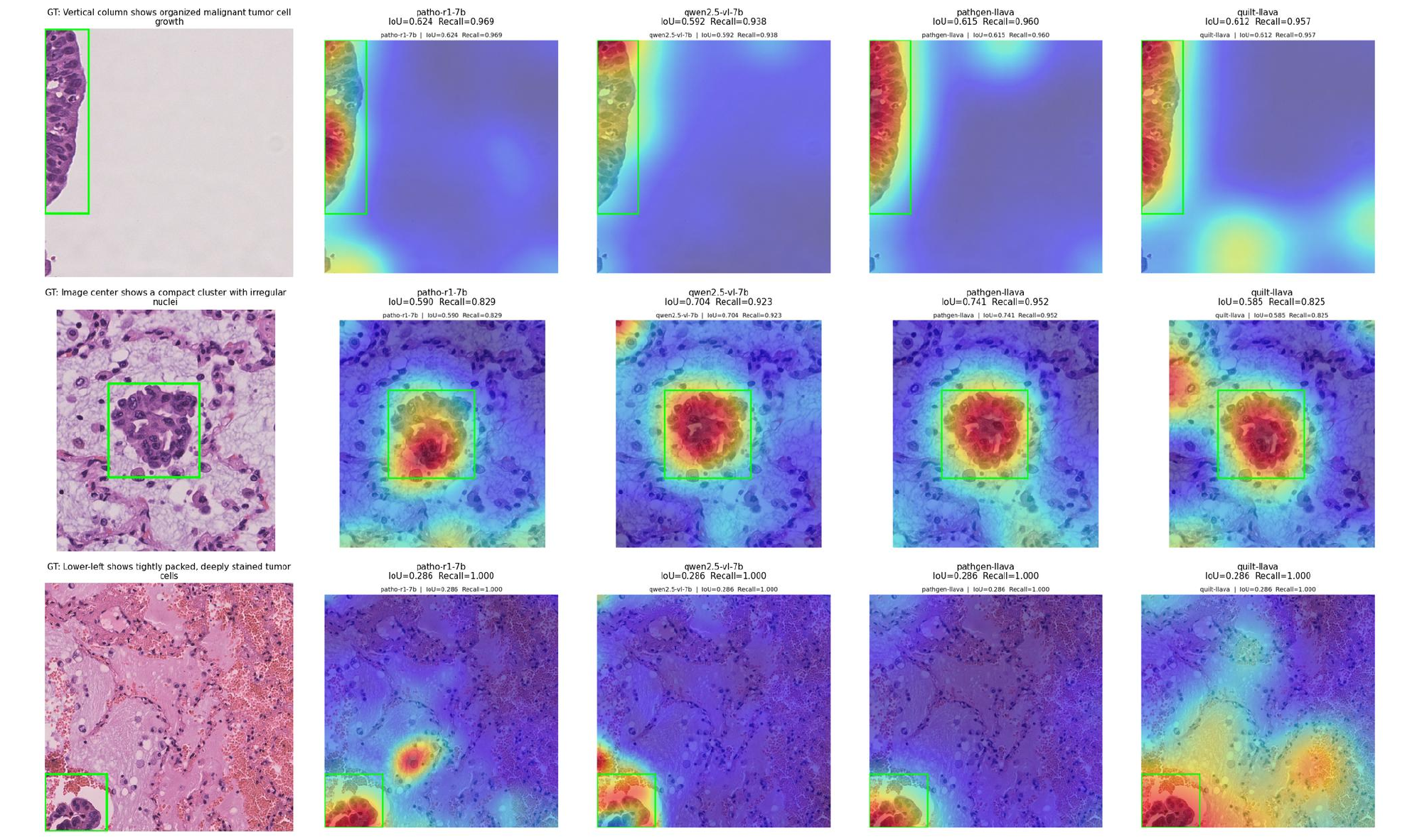}
  \caption{Successful examples of entity-token attention grounding on PathVG.}
  \label{attention_good}
\end{figure*}

\begin{figure*}[t!]
  \includegraphics[width=1.\textwidth]{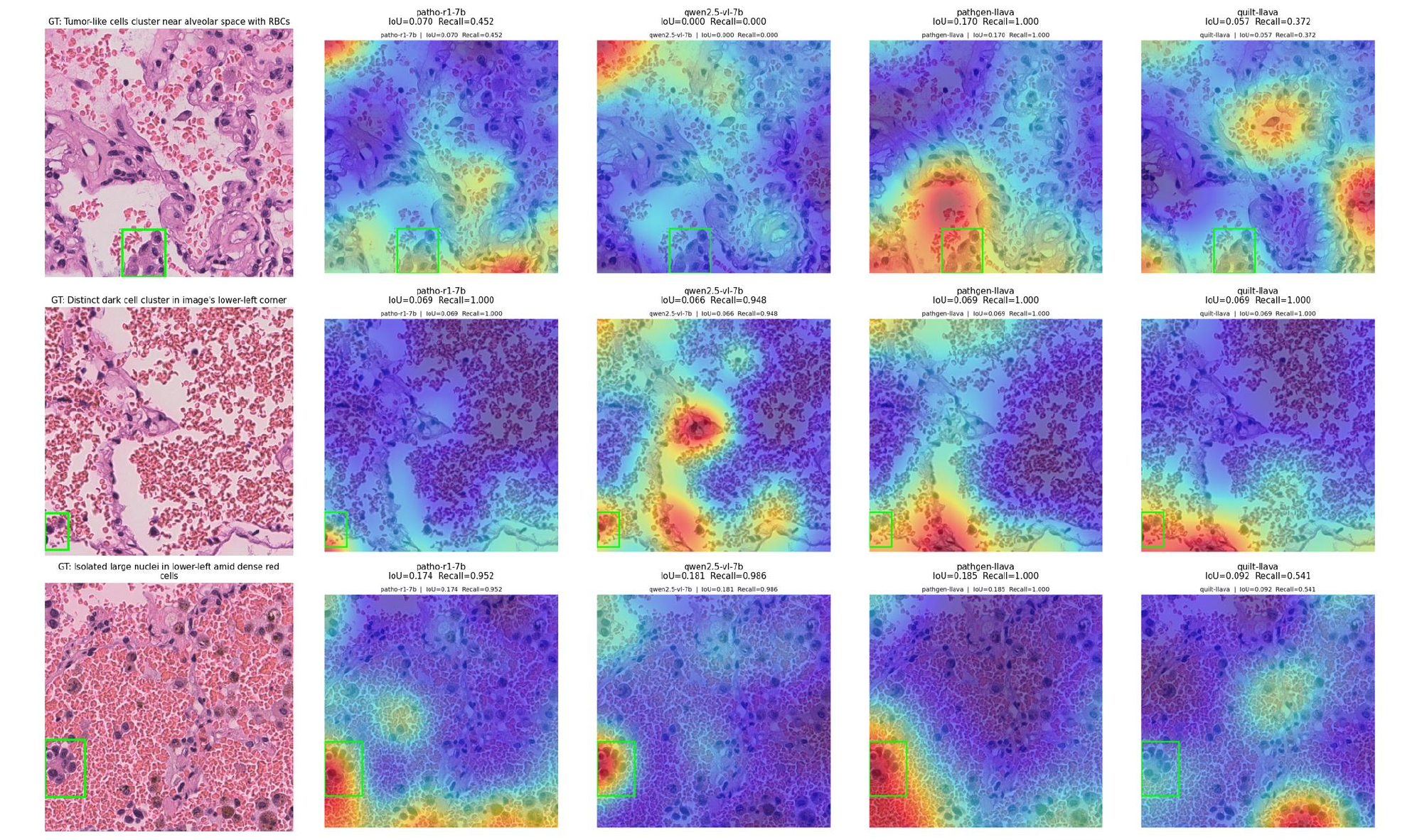}
  \caption{Partial examples of entity-token attention grounding on PathVG.}
  \label{attention_mid}
\end{figure*}

\begin{figure*}[t!]
  \includegraphics[width=1.\textwidth]{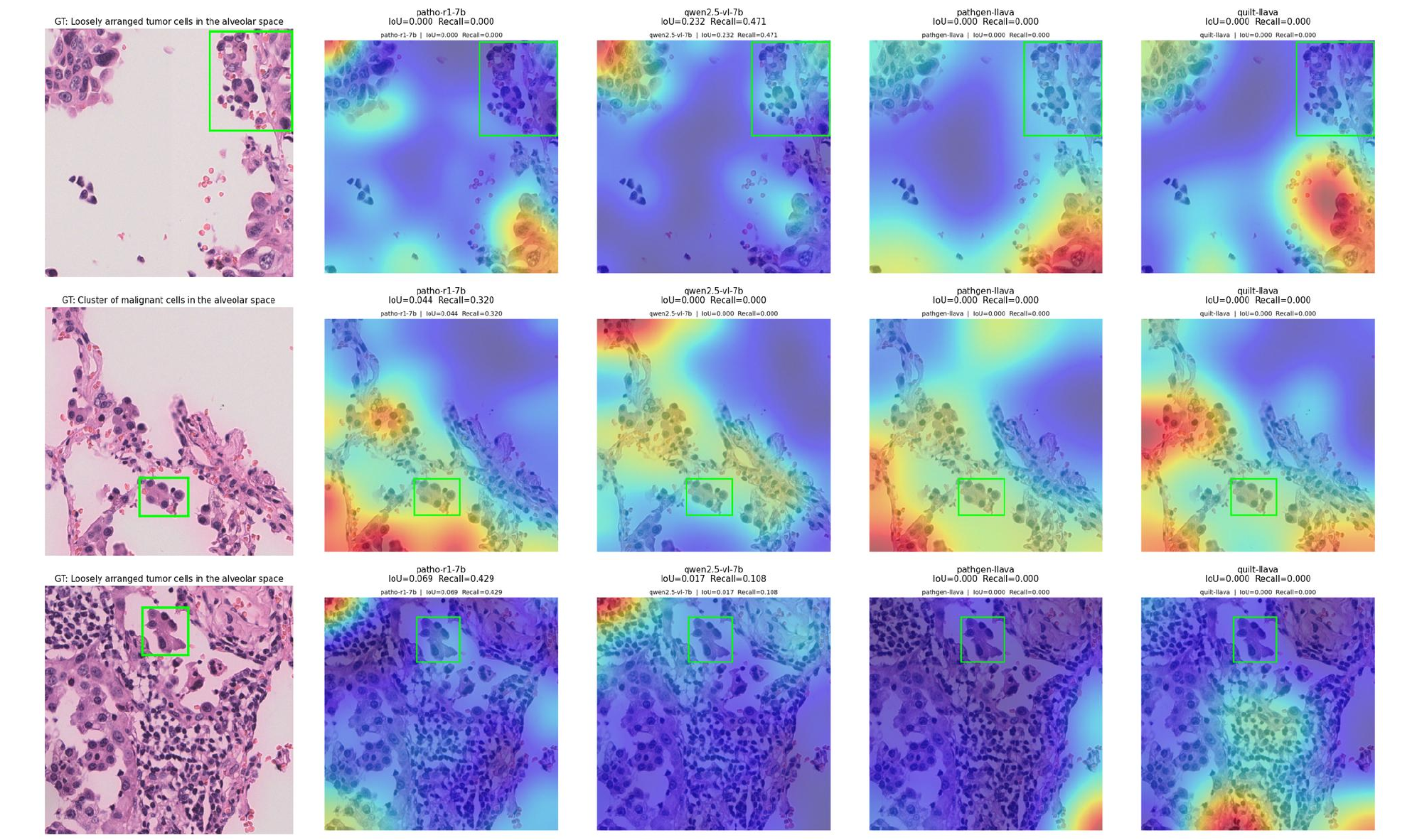}
  \caption{Failure examples of entity-token attention grounding on PathVG.}
  \label{attention_fail}
\end{figure*}

Figures \ref{attention_good}-\ref{attention_fail} provide qualitative examples of entity-token attention grounding on PathVG. We group examples into three representative categories: successful localization, partial localization, and failed localization. These visualizations complement the quantitative results in Table \ref{table A3} and illustrate the different forms of attention-grounding behavior observed across models.

In successful cases, shown in Figure \ref{attention_good}, the attention heatmaps substantially overlap with the annotated pathological regions. In partial cases, shown in Figure \ref{attention_mid}, attention may touch the target region but remains spatially diffuse, shifted, or mixed with nearby structures. In failure cases, shown in Figure \ref{attention_fail}, attention is largely misaligned with the ground-truth box or focuses on irrelevant regions.

These examples help explain why recall can be relatively high while IoU and precision remain low. Many models partially cover the correct region but also spread attention over broad irrelevant areas. This supports our observation that pathology VLMs may loosely attend to relevant visual content without forming precise entity grounding.

In Figures \ref{attention_good}-\ref{attention_fail} some attention maps emphasize boundary regions rather than the semantic core of the annotated pathological structure. This pattern can arise from both a shared extraction choice and architecture-specific visual tokenization. First, Gaussian smoothing in Algorithm \ref{alg:attention_extraction}, which is applied to all models, spreads attention responses from the entity core to adjacent patches, particularly when the annotated region is small relative to the smoothing kernel. Second, for Qwen-VL models, spatial token merging before decoding produces a coarser patch grid whose structure is retained during nearest-neighbor upsampling. Thus, these boundary effects need not share the same source across model families. The threshold sweep in Table~\ref{tab:sweep_thr} preserves the ranking of the six evaluated models, suggesting that these effects do not drive the rank-based comparison for this subset.

\subsection{On interpreting absolute and pair-difference IoU}
\label{sec:C6}
Several evaluated models, most notably Qwen2.5-VL, Qwen3-VL and InternVL3, are pretrained with explicit bounding-box and spatial grounding objectives, whereas the LLaVA-1.5 family and Quilt-LLaVA are trained primarily on image-text pairs without spatial supervision. This asymmetry gives spatially supervised models a structural advantage on any region-level attention metric, and we do not claim that models without spatial supervision should match spatially supervised models in absolute IoU; the absolute rows of Table~\ref{table3} are provided for completeness.

Our load-bearing claim is instead the pair-difference analysis of Figure~\ref{fig4}. Within each pair, both models inherit the same base-level spatial supervision, so $\Delta$IoU and $\Delta$Recall isolate the additional visual-grounding effect associated with pathology-specific tuning.

\section{Prompt Templates}
\label{appD}
This section provides the prompt templates used in our experiments. We use separate prompts for multiple-choice VQA, yes/no VQA, entity-attention extraction, and grounding-expression filtering. The multiple-choice and yes/no prompts are used for all experiments involving the corresponding datasets, including the VQA evaluations in the main paper, the extended text-only evaluations in the appendix, image perturbation experiments, and the VQA filtering stage of PathBind construction. For non-thinking models, we remove the instruction requiring the model to output reasoning inside the tags and retain only the structured final-answer format.

\subsection{Prompts for VQA Evaluation}
\label{appD1}
\begin{figure*}[t!]
  \includegraphics[width=1.\textwidth]{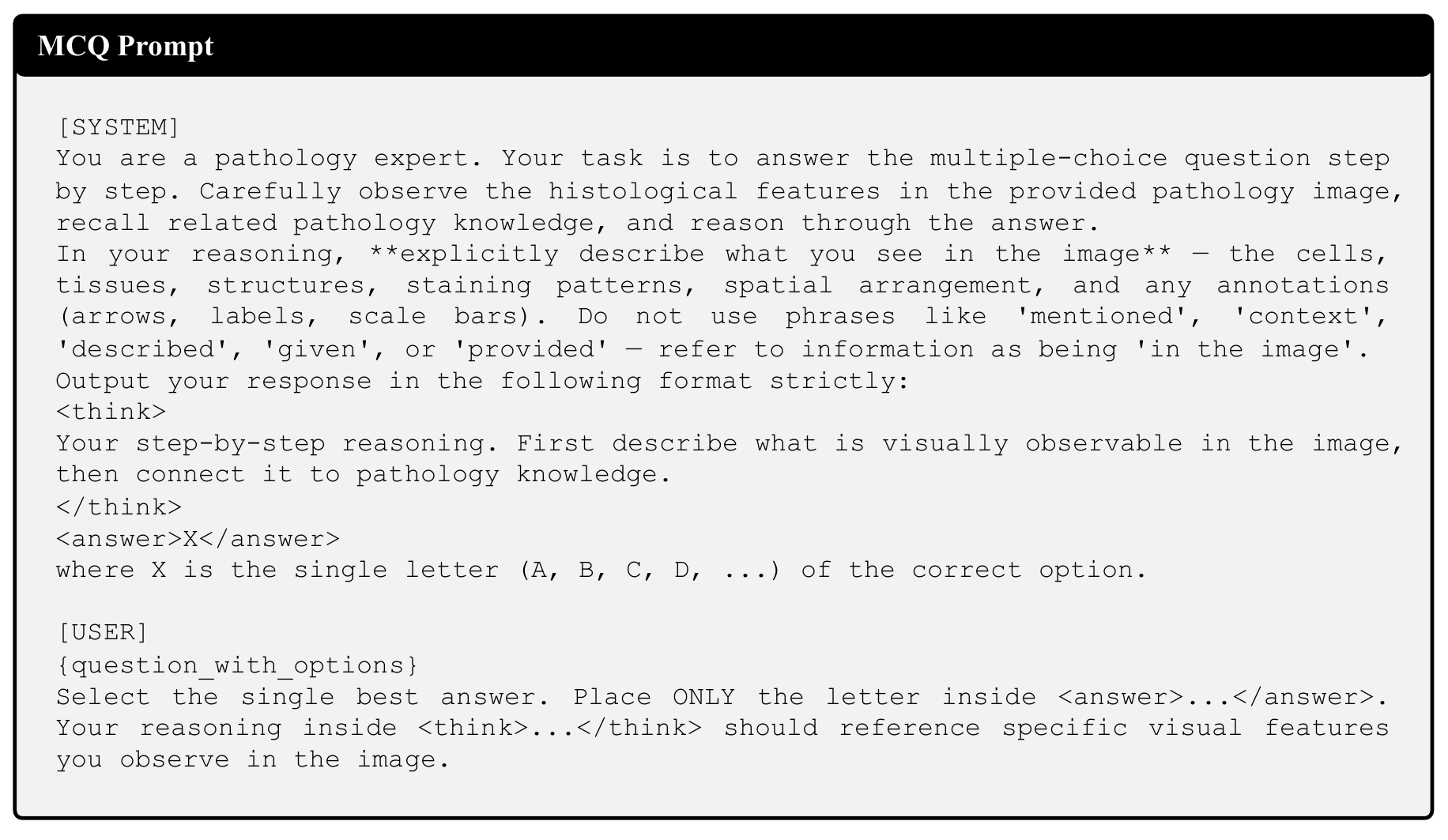}
  \caption{Prompt templates used for MCQ datasets, including PathMMU, MedXpert-Path, and OmniMed-Bright.}
  \label{mcq_prompt}
\end{figure*}

For multiple-choice datasets, including PathMMU, MedXpert-Path, and OmniMed-Bright, we use the prompt template shown in Figure \ref{mcq_prompt}. The prompt instructs the model to examine histological visual evidence, reason over the question, and output the final option in a structured field. The fixed answer format allows consistent parsing across models and evaluation settings.

\begin{figure*}[t!]
  \includegraphics[width=1.\textwidth]{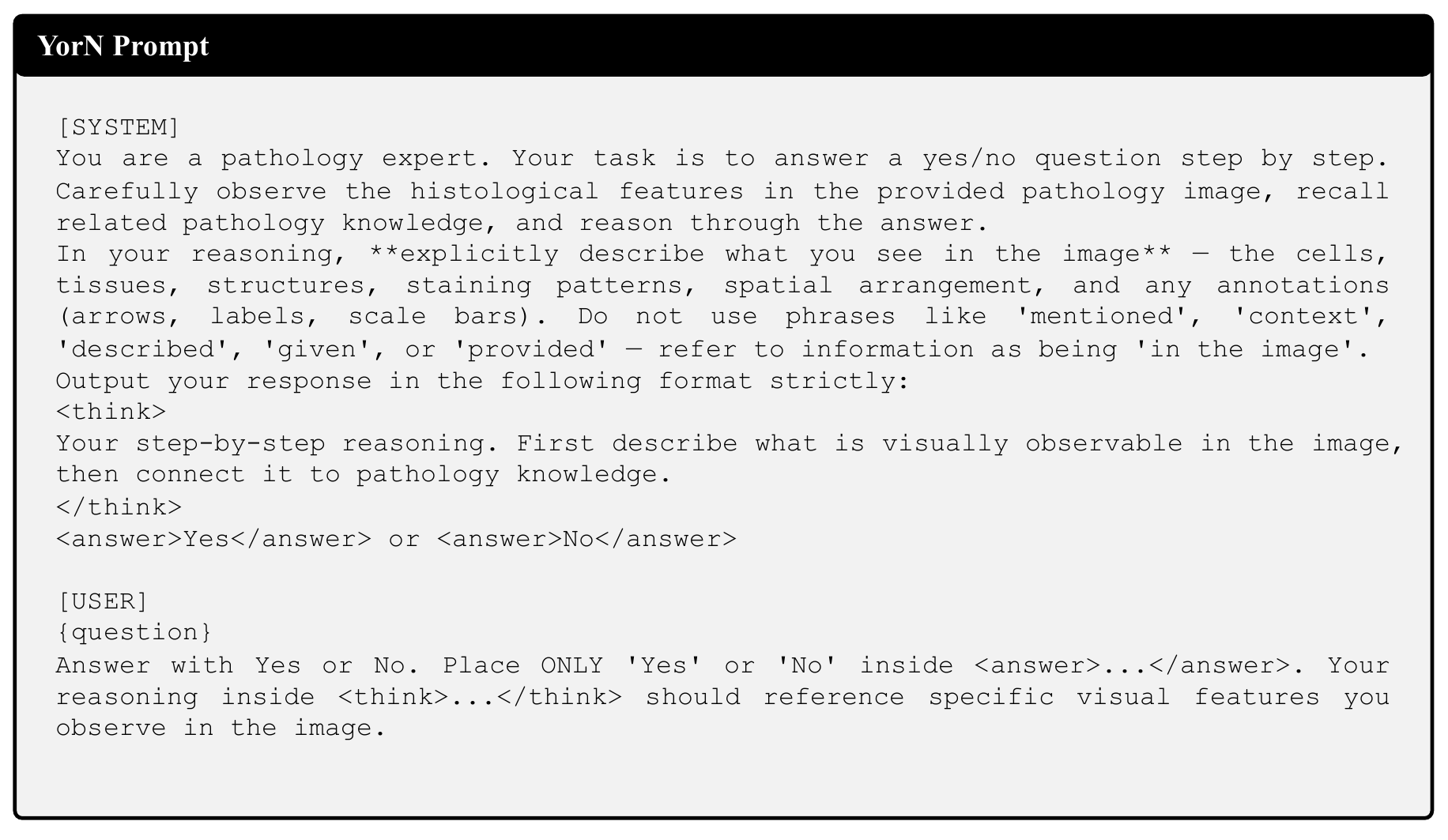}
  \caption{Prompt templates used for Yes or No datasets, including Quilt-VQA and Path-VQA.}
  \label{yorn_prompt}
\end{figure*}

For yes/no datasets, including Quilt-VQA and Path-VQA, we use the prompt template shown in Figure \ref{yorn_prompt}. The prompt follows the same design principle as the multiple-choice prompt, but constrains the final answer to either \texttt{Yes} or \texttt{No}. In text-only experiments and VQA filtering, the image input is removed while the same question prompt and answer format are preserved.

\subsection{Prompt for Entity-Attention Extraction}
\label{appD2}
For attention-based grounding analysis, we use a minimal localization prompt:
\texttt{Look at the image and locate: \{entity\}}
where \texttt{{entity}} is replaced by the pathological expression associated with the ground-truth region. This prompt is not used to evaluate answer correctness. Instead, it is designed to expose the entity phrase in the input sequence so that we can extract the attention from the corresponding entity tokens to image patches. The extracted attention maps are then used for the grounding metrics described in Appendix \ref{appC}.

\subsection{Prompt for Grounding Expression Filtering}
\label{appD3}
\begin{figure*}[t!]
  \includegraphics[width=1.\textwidth]{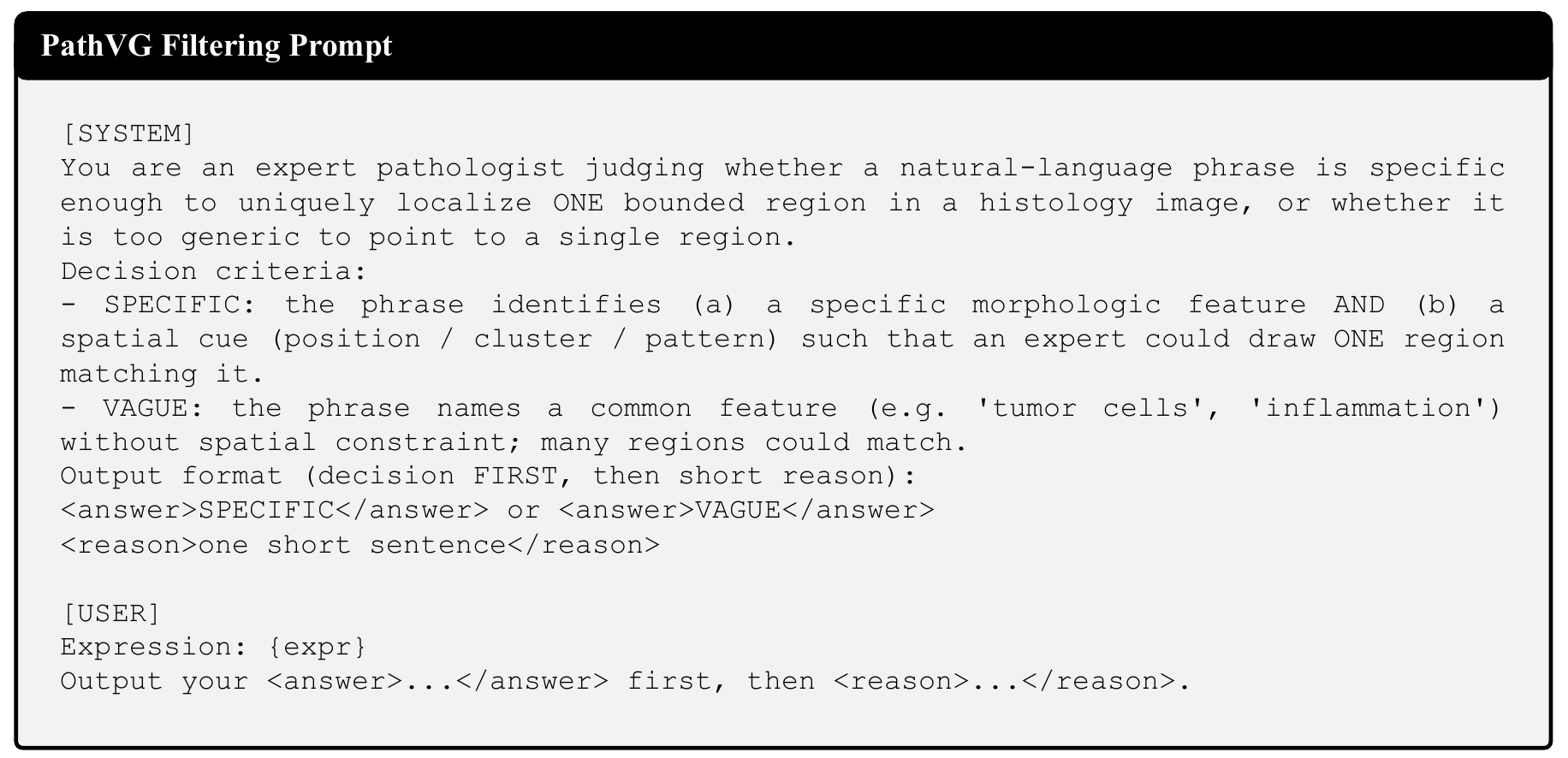}
  \caption{Prompt templates used for filtering the PathVG dataset.}
  \label{pathvg_filter_prompt}
\end{figure*}

For constructing the grounding component of PathBind, we filter grounding expressions according to whether they are specific enough to localize one bounded region in a histology image. The prompt template is shown in Figure~\ref{pathvg_filter_prompt}. The model is asked to classify each expression as \textsc{Specific} or \textsc{Vague}. Expressions classified as \textsc{Specific} are retained, while \textsc{Vague} expressions are removed.

\subsection{Private VQA Generation Prompt}
\label{appD4}
\begin{figure*}[t!]
  \includegraphics[width=1.\textwidth]{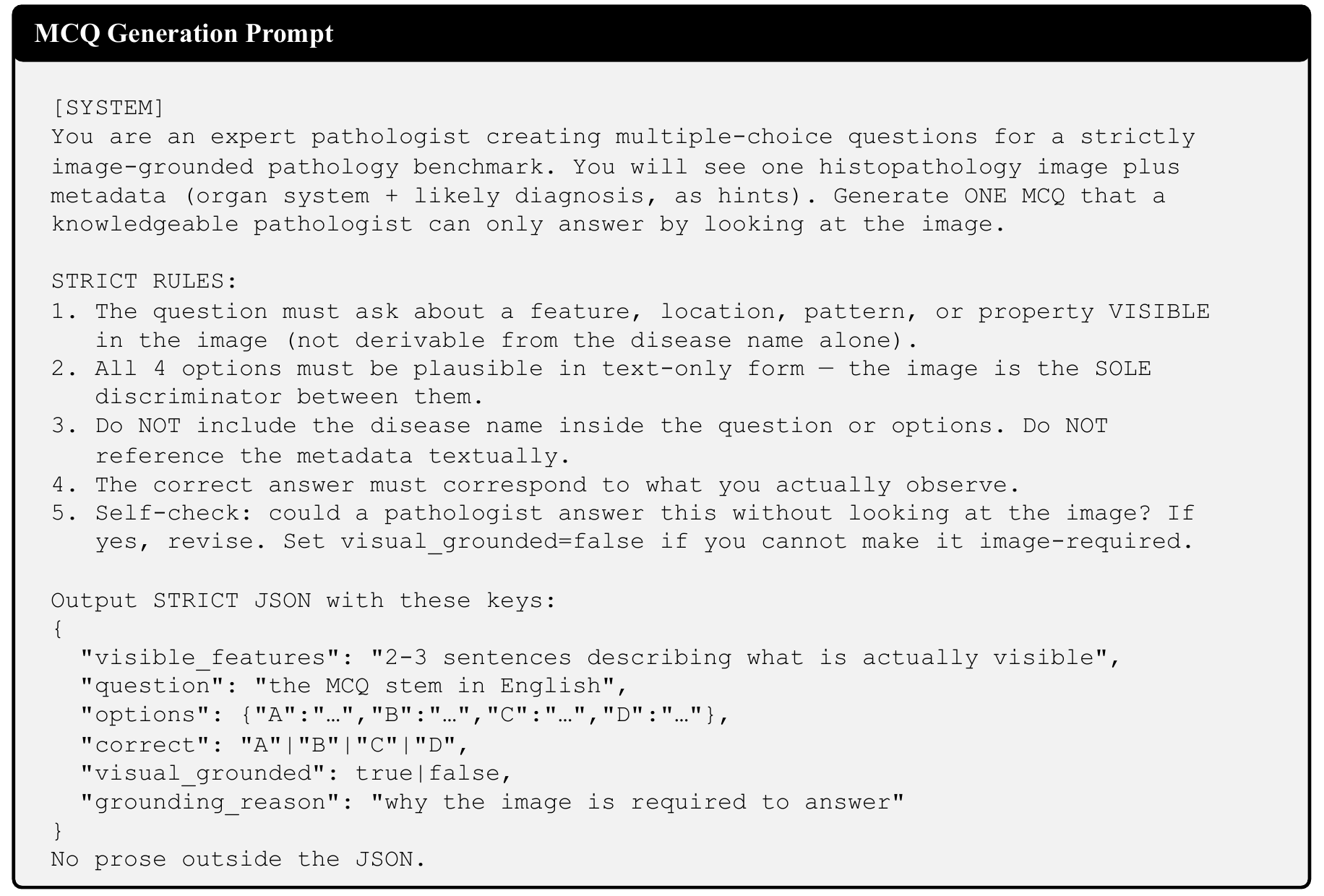}
  \caption{Prompt templates used for PathBind-PTA generation.}
  \label{private_vqa_generation_prompt}
\end{figure*}

To construct the private VQA component of PathBind, we use GPT-5.4 to generate initial multiple-choice pathology VQA samples from selected pathology teaching-atlas images. The prompt template used for image-based question generation is shown in Figure~\ref{private_vqa_generation_prompt}. The generated samples are subsequently reviewed by pathology experts to remove ambiguous cases, text-leakage samples, and questions that lack sufficient visual evidence.

\section{Implementation Details}
\label{appE}
All experiments are conducted under fixed inference configurations unless otherwise specified. Closed-source models, including GPT-5.4 and Gemini-3-Pro, are evaluated through official API services. The same API-based inference protocol is used for the LLM/VLM models involved in the automated filtering stage of PathBind-VQA.

All open-source models are evaluated locally on a workstation equipped with eight NVIDIA RTX 4090 GPUs using publicly released checkpoints. For all VQA experiments, we use consistent prompts and answer parsing procedures across models. In the text-only setting, only the image input is removed while keeping the instruction, question, and output format unchanged. Model outputs are generated with fixed decoding configurations.

For attention-based grounding analysis, we only evaluate models with publicly accessible attention weights. Entity-token attention maps are extracted and aggregated across layers and heads, followed by the grounding metric computation described in Appendix \ref{appC1}-\ref{appC3}.


\end{document}